\documentclass[10pt,twocolumn,letterpaper]{article}

\usepackage[pagenumbers]{cvpr} 

\usepackage[dvipsnames]{xcolor}
\definecolor{cvprblue}{RGB}{0,0,255}
\definecolor{darkgreen}{RGB}{0,127,0}
\definecolor{darkred}{RGB}{200,0,0}
\usepackage{multirow}
\usepackage{verbatim}
\usepackage{graphicx}
\usepackage{booktabs}
\usepackage{xcolor}
\usepackage{makecell}
\usepackage{bigstrut}
\usepackage{hhline}
\usepackage{pifont}  
\usepackage{xspace}
\usepackage[table]{xcolor}
\usepackage{booktabs}
\usepackage{bigstrut}
\usepackage{placeins}
\usepackage{gensymb}
\usepackage{soul}
\usepackage{wrapfig,lipsum}
\usepackage{tikz}
\usepackage{yfonts}
\usepackage{import}
\usepackage{gensymb}
\usepackage{geometry}
\usepackage{array}
\usepackage{placeins}
\usepackage{longtable}

\makeatletter  
\@namedef{ver@everyshi.sty}{}

\def\greencheckmark{\textcolor{darkgreen}{\checkmark}}
\def\redxmark{\textcolor{darkred}{\ding{55}}}

\newcommand{\thickline}[0]{\Xhline{3pt}}

\makeatletter
\newcommand\blfootnote[1]{%
  \begingroup
  \renewcommand\thefootnote{}\footnote{#1}%
  \addtocounter{footnote}{-1}%
  \endgroup
}
\makeatother

\definecolor{cvprblue}{rgb}{0.21,0.49,0.74}
\usepackage[pagebackref,breaklinks,colorlinks,citecolor=cvprblue]{hyperref}

\definecolor{blue}{rgb}{0.22, 0.22, 0.95}

\def\paperID{11293} 
\def\confName{CVPR}
\def\confYear{2025}

\author{
Minjun Kang\textsuperscript{1}$^{\ast}$ \quad
Inkyu Shin\textsuperscript{2}$^{\ast\dagger}$ \quad
Taeyeop Lee\textsuperscript{1} \quad
In So Kweon\textsuperscript{1} \quad
Kuk-Jin Yoon\textsuperscript{1} \\
\textsuperscript{1}KAIST \quad
\textsuperscript{2}ByteDance Seed \\
}

\begin{document}


\title{Drag4D: Align Your Motion with Text-Driven 3D Scene Generation}

\maketitlewithvisual{We propose Drag4D, a comprehensive user-interactive framework 
for 4D-controllable video generation, designed to achieve spatial and temporal
alignment of a target instance within a text-driven 3D background. For example, this framework allows users to create a high-quality 3D scene from a text description
(middle section), seamlessly integrate target instances (left section), and precisely control motion following a user-defined 3D trajectory (right section).}

\blfootnote{$^{\ast}$ Equal contribution.}
\blfootnote{$\dagger$ This work was partially conducted while at KAIST.}

\begin{abstract}
We introduce \textbf{Drag4D}, an interactive framework that integrates object motion control within text-driven 3D scene generation. This framework enables user to define 3D trajectory for the 360\degree objects generated from a single image, seamlessly integrating them into a high-quality 3D background. Our Drag4D pipeline consists of three stages. 
First, we enhance text-to-3D background generation by applying 2D Gaussian Splatting with panoramic images and inpainted novel views, resulting in dense and visually complete 3D reconstructions.
In the second stage, given a reference image of the target object, we introduce a 3D copy-and-paste approach: the target instance is extracted in a full 360° representation using an off-the-shelf image-to-3D model and seamlessly composited into the generated 3D scene. The object mesh is then positioned within the 3D scene via our physics-aware object position learning, ensuring precise spatial alignment. Lastly, the spatially aligned object is temporally animated along a user-defined 3D trajectory. To mitigate motion hallucination and ensure view-consistent temporal alignment, we develop a part-augmented, motion-conditioned video diffusion model that processes multiview image pairs together with their projected 2D trajectories.
We demonstrate the effectiveness of our unified architecture through evaluations at each stage and in the final results, showcasing the harmonized alignment of user-controlled object motion within high-quality 3D background.
\end{abstract}

\section{Introduction}
\label{sec:intro}

The importance of user experience in computer vision has grown significantly. It has been driven by the rapid advancement of applications in various domains, specifically in VR/AR~\cite{rambach2021survey}. Enhancing user experience lies the ability to control and manipulate digital environments intuitively and interactively. One area where this control is particularly transformative is in video content, where users seek to generate their own videos with text descriptions~\cite{blattmann2023stable, liu2024sora, polyak2024movie, jin2024pyramidal}. Due to the ambiguity of user-intention in text prompt, video model with user-given trajectory~\cite{jain2024peekaboo, wu2025draganything} has recently emerged to allow users to effectively direct and modify the objects within videos. The another area is 3D content generation~\cite{poole2022dreamfusion, shi2023MVDream, hoellein2023text2room, zhou2024holodreamer, li2024scenedreamer360, zhou2024layout}. It can construct 3D space, enabling users to engage with generating objects~\cite{poole2022dreamfusion, shi2023MVDream} or/and scenes~\cite{hoellein2023text2room, zhou2024holodreamer, li2024scenedreamer360} in a fully immersive manner with unconstrained multiple camera views. The module designed for controllable video generation with trajectory and text-to-3D generation are not only evolving over time, but also becoming very distinct. Consequently, it is infeasible to easily adapt temporal controllability to text-to-3D (and vice versa). Particularly, current 2D trajectory-based video generation~\cite{jain2024peekaboo, wu2025draganything} lacks the capability to scale up to multi-views, while text-to-3D method~\cite{chung2023luciddreamer, li2024scenedreamer360, zhou2024layout} can suffer from misalignment of user-given moving objects with generated 3D scenes as shown in~\cref{table:components}. The need for this scenario-specific design results in degrading the quality of user experience. A natural question thus emerges: \textit{Is it feasible to develop a unified framework capable of achieving controllable 4D environments, where user can manipulate the trajectory of 360\degree object while synthesizing 3D background?}

\begin{table}[t!]
\captionof{table}{Our approach, Drag4D, seamlessly unifies three key components required for 4D controllable video generation: (1) generating a 3D scene from a text prompt, (2) composing objects into the generated 3D background, and (3) conditioning video motion to manipulate an object's trajectory naturally within the 3D space. No prior work has achieved this level of integration.}
\centering
\definecolor{grey}{RGB}{230,230,230}
\scalebox{0.5}{
\setlength{\tabcolsep}{3pt} 
\begin{tabular}{l|cccc}
\hline
Method            & 3D Scene Generation & Object Composition   & Motion Conditioned Video  \\
\hline
LucidDreamer~\cite{chung2023luciddreamer} & O & X & X \\
SceneDreamer360~\cite{li2024scenedreamer360} & O & X & X \\
Layout3D~\cite{zhou2024layout} & X & O & X \\
DragAnything~\cite{wu2025draganything} & X & X & O (2D Video) \\
\hline
Drag4D & O & O & O (4D Video) \\
\hline

\end{tabular}}
\label{table:components}
\end{table}

To answer the question, we present a unified user-interactive framework, \textbf{Drag4D}, which aims to align your motions with text-to-3D generation. Our pipeline incorporates three key stages.
First, in order to construct a basis for 3D background scene, we employ off-the-shelf text-driven panoramic image generation model~\cite{zhang2024taming}. 
Then, we extract depth and normal information from the panoramic image using a depth estimator~\cite{hu2024metric3d}, providing essential priors (point cloud) for 3D scene reconstruction.
In constrast to SceneDreamer360~\cite{li2024scenedreamer360, shi2023MVDream}, which directly augments training set with novel view images projected from point cloud, our approach integrates image inpainting model~\cite{esser2024scaling} to refine these novel views like HoloDreamer~\cite{zhou2024holodreamer}. This inpainting process effectively addresses occlusions from different camera views, seamlessly filling in missing regions to enhance the visual coherence in the reconstructed 3D scene. Our framework departs from HoloDreamer~\cite{zhou2024holodreamer} by employing efficient 2D Gaussian Splatting~\cite{huang20242d} to be optimized jointly with panoramic image set and inpainted novel view images.
Specifically, we enhance the joint learning by proposing a pixel-level adaptive weighting mechanism using depth-normal similarity, which can mitigate the influence of noisy areas in augmented views. 
Secondly, to extract the instance from user-provided reference image and composite it into the generated 3D background, we propose a 3D copy-and-paste approach. We employ an image-to-3D model~\cite{xu2024instantmesh} to scale up the area of foreground mask to 360\degree object (``3D copy''). Then, it is spatially aligned with the surrounding 3D background through physics-aware object position learning (``3D paste''), which can be implemented via collision and gravity loss.
In the final stage, we input multiple views of the spatially composited scene-object and the corresponding trajectories (derived from a user-provided 3D path) into motion-conditioned video diffusion model. Unlike a previous model~\cite{wu2025draganything}, which controls the motion of an entire instance with a single global feature, we introduce a part-augmented motion-conditioned video generator where local feature is co-utilized with global feature. We coin this approach, \textbf{L}ocal-\textbf{G}lobal DragAnything.
We observe that this approach effectively prevents local motion hallucinations within the target instance, which is essential for precise motion alignment. 
We rigorously validate our design choices and methodology through comprehensive experiments presented in this paper. Given the limited availability of datasets specifically targeting moving objects within 3D scenes, we introduce a custom dataset, Drag4D-30, to showcase the enhanced performance of our approach compared to baseline methods.







\begin{figure*}[t!]
\centering
\includegraphics[width=\textwidth]{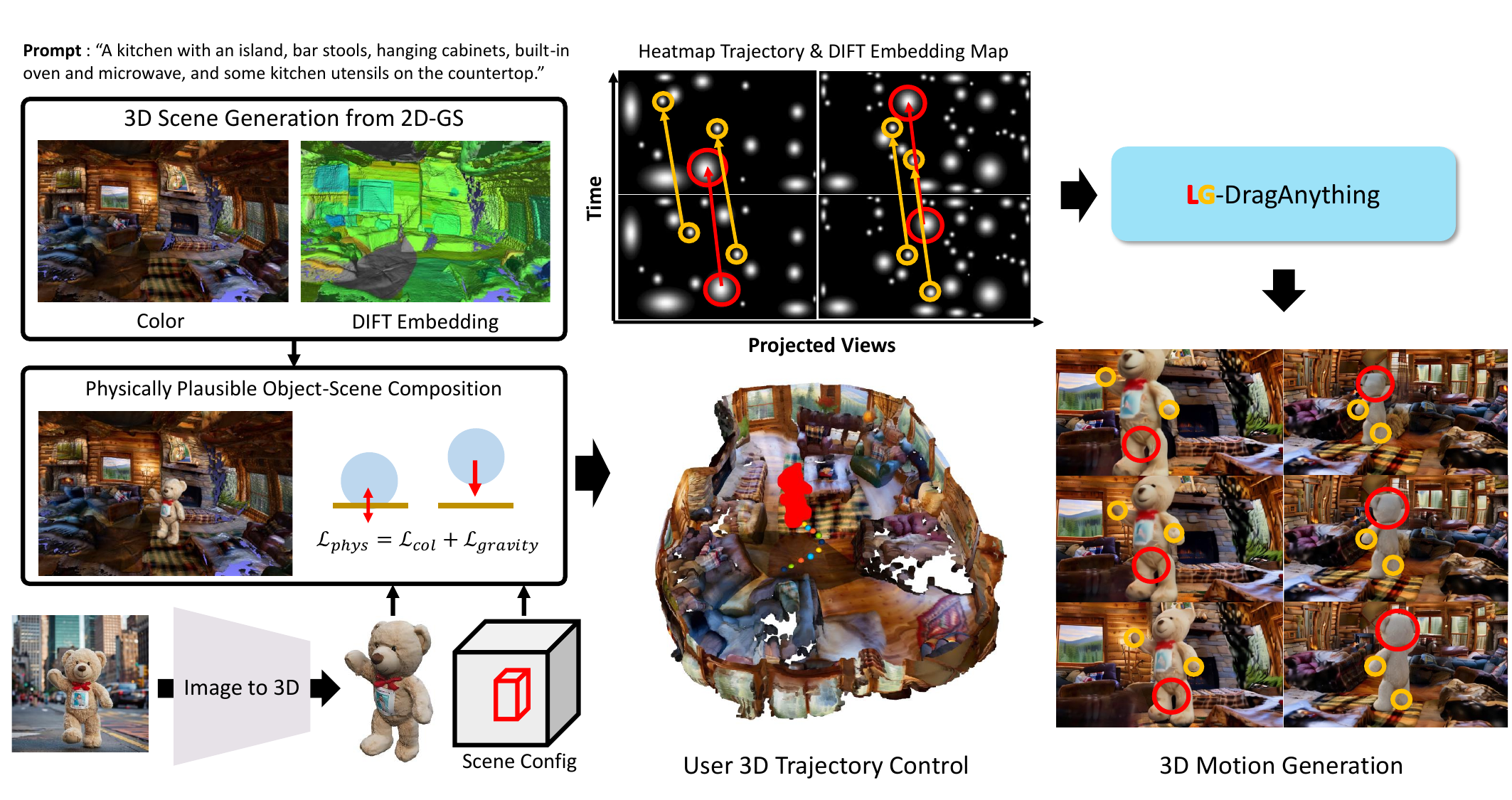}
\caption{The proposed Drag4D comprises three key stages. First, we conduct text-to-3D scene generation. Here, we use 2D Gaussian Splatting to process panoramic and inpainted augmented view images, generating high-quality 3D scenes with diffuion features (DIFT). Next, given reference image with target instance, a 360\degree object mesh is extracted from a reference image using an Image-to-3D model and composited into the 3D background based on scene configuration (e.g., 3D bounding boxes). A physics-aware object-scene composition method including collision loss and gravity loss ensures accurate spatial alignment of the target instance. In final stage, with the composited 3D scene and a user-defined 3D trajectory, the LG-DragAnything motion-conditioned video model enables view-consistent multi-view video generation, achieving high-quality 3D motion alignment.} 
\label{fig:3D_scene_generation}
\end{figure*}

\section{Method}
The meta architecture of Drag4D aims to design 4D controllable video generation, which align and manipulate object within 3D scene background generation. This process unfolds across three seamlessly integrated stages. The first stage, detailed in [\cref{sec:1st}], generates a 3D background scene by transforming given text prompt into high-quality panoramic image. In the second stage, described in [\cref{sec:2nd}], a user-defined object is extracted from a reference view in a 360\degree manner and spatially composited into the generated 3D background. Finally, the third stage, covered in [\cref{sec:3rd}], temporally aligns the user-specified 3D trajectory with the object, maintaining seamless composition with the surrounding background.

\subsection{Preliminaries}
\noindent\textbf{Diffusion Models}
Diffusion probabilistic models (DPMs), first introduced by \citep{sohl2015deep} and further refined by \citep{ho2020denoising}, constitute a type of generative model that reconstructs a target data distribution, denoted as $q$, through a staged denoising process. The process begins with an image $x_T$ that is initially Gaussian-distributed as $x_T \sim \mathcal{N}(0, I)$. Containing independent and identically distributed noise. The diffusion model, represented by $\epsilon_{\theta}$ , then progressively reduces this noise, transforming the image step-by-step until it arrives at a clean version, $x_0$ drawn from the target distribution $q$.

\noindent\textbf{3D Gaussian Splatting (3D-GS)}
Kerbl et al.~\cite{kerbl20233d} introduce a method for representing 3D scenes using 3D Gaussian primitives and rendering images through differentiable volume splatting. In this approach, 3D-GS explicitly defines Gaussian primitives by specifying their 3D covariance matrix $\Sigma$ and spatial location $p_k$:

\begin{equation}
    \mathcal{G}(\mathbf{p}) = \exp\left(-\frac{1}{2}(\mathbf{p} - \mathbf{p}_k)^\top \Sigma^{-1} (\mathbf{p} - \mathbf{p}_k)\right)
\end{equation}

Here, the covariance matrix $\Sigma$ is decomposed into a scaling matrix $S$ and a rotation matrix $R$, such that $\Sigma$ = $RSS^{\top}R^{\top}$. To render an image, the 3D Gaussian is transformed to the camera's coordinate system using a world-to-camera transformation matrix $W$, and then projected onto the image plane via a local affine transformation. This results in a modified covariance matrix:

\begin{equation}
    \Sigma' = J W \Sigma W^\top J^\top 
\end{equation}

\subsection{Problem Setting}
\label{sec:prob}
Our pipeline utilizes multiple user prompts across different stages to fully capture and reflect user's intentions. In the first stage, a detailed and long text prompt $t$ is provided to generate high-quality 3D background scene. In the subsequent stage, a single reference image, $x_r$ with an foreground mask ($m_r$) is supplied to extract the target object, which is then transformed into 360\degree object. Afterwards, it is scaled and positioned within generated 3D background scene according to a user-defined 3D bounding box configuration, $B_{3D}$ , which specifies center coordinates, dimensions and rotation. Finally, user can manipulate the object within the 3D space using a sequence of 3D trajectory points, $(p_x,p_y, p_z)_{i=1...N}$, where $N$ denotes the number of points defining the trajectory path.

\subsection{1st Stage: Generating Text-to-3D Scene}
\label{sec:1st}

\noindent\textbf{Panoramic Image Generation for 3D Scene} The objective of the panoramic image generator is to construct high-quality 360-degree scenes guided by text input $t$, serving as a critical prior for reconstructing 3D backgrounds. To this end, we employ a diffusion model~\cite{zhang2024taming} for generating panoramic images.
To effectively handle extended text prompts and generate high-resolution panoramic images, we incorporate LoRA~\cite{hu2021lora} fine-tuning and super-resolution~\cite{wang2018esrgan} techniques, following SceneDreamer360~\cite{li2024scenedreamer360}. Then, we can obtain the panoramic image $I_p$ along with its corresponding depth $D_p$ using a pre-trained metric depth estimator, Metric3D~\cite{yin2023metric3d}.
Given panoramic image $I_p$ with corresponding depth $D_p$, we can obtain point cloud, $P$, utilizing \footnote{Mapping function to transform 2D pixel coordinates of panorama image into 3D coordinates.}{inverse equirectangular projection}, $E^{-1}$ as below:

\begin{equation}
    P = E^{-1}(I_p, D_p)
    \label{eqn:pointcloud}
\end{equation}
, which serves as the initialization for reconstructing a dense 3D background. To obtain a series of perspective images used as supervision for 3D reconstruction, we first can derive a high-quality set of base images by projecting from point cloud $P$ using cameras positioned at the center of the panoramic sphere. Specifically, using shared intrinsic $K$ and multiple center-positioned extrinsics $E_i$, base images can be obtained with following equation:

\begin{equation}
    I_{i} = \Phi(P, K, E_i)
    \label{eqn:projection}
\end{equation}
Let $\Phi$ denote the projection function from 3D point cloud to the corresponding 2D pixel coordinate.

\noindent\textbf{Image Inpainting for Novel Views}
However, 3D reconstruction with the supervision of the base images $I_i$ may result in poor rendering quality due to the limited range of camera poses. In our Drag4D scenario, where a dense 3D background is essential, this limitation can lead to the emergence of significant visual artifacts.
To address this, we augment extrinsics ($E_i$) of the base images, adjusting camera positions away from the center of the panoramic sphere as follows:

\begin{equation}
    \text{Aug}(I_{i}) = [\Phi(P, K, E_{ij})]_{j=1...T}
    \label{eqn:augmented}
\end{equation}
, which corresponds $T$ number of augmented views from $i$th base image. Yet, we observe significant artifacts near object boundaries in the augmented views due to depth instability. Thus, we filter out these areas based on a depth gradient threshold and fill them in using pretrained stable diffusion model, $SD$~\cite{esser2024scaling}. The process of image inpainting for augmented views are as follows:
\begin{equation}
    \text{PaintAug}(I_{i})_{j} = SD(\Phi(P, K, E_{ij}), M_{ij})
    \label{eqn:paintaug}
\end{equation}
Here, $M_{ij}$ stems from mask filtered out from depth gradient threshold. We can then obtain a pair of base images and inpainting augmented images, $[I_{i}, \text{PaintAug}(I_i)_{j}]$.

\vspace{1mm}
\noindent\textbf{One-stage 2D-GS Optimization}
For improved 3D reconstruction, we replace 3D Gaussian primitives with 2D primitives~\cite{huang20242d}. It has been demonstrated that 2D-GS provides faster and more consistent multi-view consistency evaluations than 3D-GS, a crucial for efficient and accurate 3D reconstruction. It is easily implemented by skipping the third row and column of $\Sigma'$ and deriving normal primitive from orthogonal of two tangential vectors. Following the training procedure of 2D-GS~\cite{huang20242d}, we can optimize our model from an initial sparse point cloud $P$ using our panoramic base image $I_{i}$ with following objective:
\begin{equation}
    L_{\text{base}} = L(G_{\theta}(P, K, E_i), I_i)
    \label{eqn:gs_render}
\end{equation}
Here, $L$ represents an integration of the reconstruction loss~\cite{kerbl20233d}, two regularizers~\cite{huang20242d} (e.g., depth distortion loss and depth-normal consistency loss). $G_{\theta}$ denotes Gaussian model with $\theta$ parameter.
Furthermore, to fully leverage the inpainted augmented views $\text{PaintAug}(I_i)_j$ while preventing the model from being constrained by noisy regions, we apply depth-normal similarity as a weighting factor for the inpainted areas of $\text{PaintAug}(I_i)_j$. We simplify the corresponding equation using rendered image of augmented view $R_{\text{aug}} = G_{\theta}(P, K, E_{ij})$
as below:
\begin{equation}
\small
\begin{split}
    L_{\text{aug}} = L((1-M_{ij})R_{\text{aug}} \text{+}
    C_{ij}{M_{ij}R_{\text{aug}}}, \text{PaintAug}(I_i)_j)
    \label{eqn:gs_aug}
\end{split}
\end{equation}
$C_{ij}$ denotes depth-normal similarity value. 

Finally, we embed semantic features including DINO~\cite{oquab2023dinov2} and DIFT~\cite{tang2023emergent} in 3D geometry to be utilized as prior for semantic-level motion-based video generation in Stage3 of~\cref{sec:3rd}.
Inspired by 3DitScene~\cite{zhang20243ditscene} and LangSplat~\cite{qin2024langsplat}, we apply feature distillation loss, $L_{\text{distill}}$, between rasterized features of the Gaussian Splats and the semantic features constrained by the SAM2~\cite{ravi2024sam2} mask, both on base and augmented images.
Therefore, total objective loss is expressed as following:
\begin{equation}
    L_{\text{scene}} = L_{\text{base}} + L_{\text{aug}} +  L_{\text{distill}}
    \label{eqn:1st_gs_loss}
\end{equation}
We skip the summation of loss for simplicity.
This approach allows for joint optimization of 2D-GS on base images and augmented images with learning 3D geometry, which contrasts with HoloDreamer~\cite{zhou2024holodreamer}, where complex multi-stage 3DGS optimization process was introduced. 
We show that our method can reconstruct an accurate 3D scene in \cref{fig:stage1_qual}.

\begin{figure}[t!]
\centering
\includegraphics[width=1.0\linewidth]{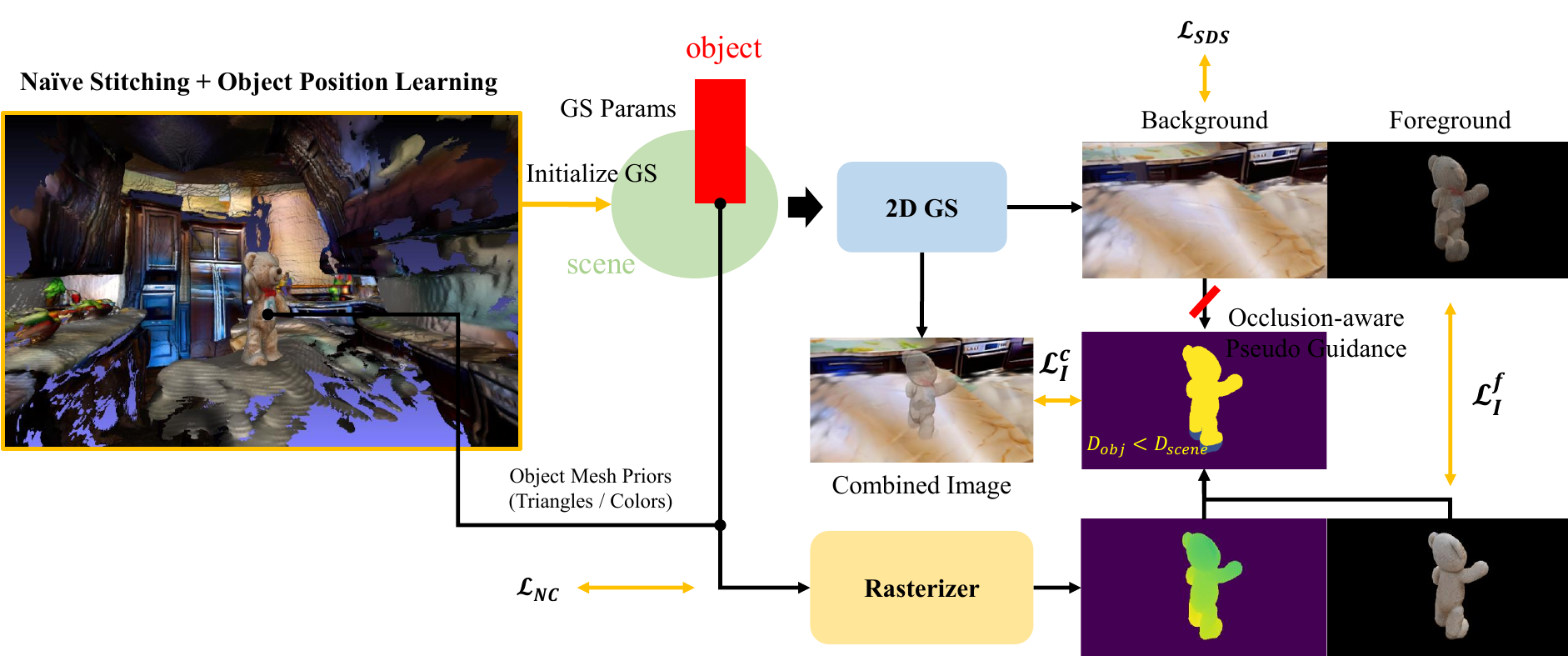}
\caption{Our object-scene composition pipeline. It begins with naive stitching between 3D background and the position-learned object mesh, which serves as an initialization for 2D-GS optimization. To refine the composition, we apply a photometric reconstruction loss to learn the opacity and spherical harmonic (SH) coefficients of the foreground object, while the background is optimized using the SDS loss~\cite{poole2022dreamfusion}.}
\label{fig:composition_method}
\end{figure}

\subsection{2nd Stage: Spatial Alignment of Object}
\label{sec:2nd}
The second stage begins with a user-provided reference image. It extracts the target instance in 3D from this image and integrates it with the generated scene from 1st stage using our proposed 3D copy-and-paste approach.
For 3D copy, we first extract the target instance from a reference image with a foreground mask. Then, we leverage off-the-shelf instance-to-3D model~\cite{xu2024instantmesh} to generate a full 3D object of the target instance. Specifically, this model use a multi-view diffusion model~\cite{shi2023zero123++} to synthesize six novel views at fixed camera poses. These generated multi-view images are then fed into a transformer-based sparse-view reconstruction model to create a high-quality 3D mesh.
Next, to perform 3D paste accurately, we take two sequential learning stages. First, we find the floor plane of the scene and roughly locate our object on the plane. We can observe that this naive way of stitching often results in unreliable poses of object as shown in the first column of \cref{fig:composition1}. 
To achieve accurate placement of composited objects, we design a physics-aware object-scene composition framework with two regularizers: 1) collision loss, which minimize the collision area between object and scene, and 2) gravity loss, which enforce the object to be grounded within the scene. Our loss term for this process is as follows:
\begin{align}
L_{\text{physics}} &= L_{\text{collision}} + L_{\text{gravity}} \nonumber \\
&= \sum_{p \in O} \sum_{q \in S} \left(1 - n_p n_q\right) \left(\|p - q\|_2 < 1e^{-3}\right) \nonumber \\
&\quad + \sum_{p \in O} \frac{1}{2} g m_p \left(p - y_{\text{floor}}\right)
\label{eq:total_loss}
\end{align}
Here, $O$ is the point cloud of the object, $S$ is the point cloud of the scene, $g$ is gravitational acceleration, and $m_p$ is the vertex mass of the object. Gravity loss penalized the gravity positional energy relative to the closest floor plane of the scene. Collision loss enforces normal consistency between points that come into contact within a distance below the threshold. This optimization stage yields well-aligned object–scene composited point clouds.

In the next stage, we initialize 2D-GS with the well-aligned object–scene composited point clouds and jointly optimize the integrated object–scene representation, as illustrated in \cref{fig:composition_method}. Consistent with the first stage, semantic features are also embedded into 2D-GS. Specifically, we reuse the semantic features of the background scene and introduce new features for the foreground object. For the latter, DINO features are employed to provide a part-level semantic prior of the target instance, thereby enabling part-level control in stage 3.
While optimizing 2D-GS of the object and the scene together, we use SDS loss~\cite{poole2022dreamfusion} $L_{\text{SDS}}$, to seemingly generate the occluded area of the background. Our total loss to optimize the scene-object composited 2D-GS is as follows:
\begin{equation}
    L_{\text{scene-object}} = L_{\text{base}} +  L_{\text{distill}} + \lambda L_{\text{SDS}}
    \label{eqn:2nd_gs_loss}
\end{equation}
Here, we use $\lambda$ as 0.01.


\begin{table*}[!t]
\caption{Comparison of Methods on Various Quality Metrics on \textit{Drag4D-30} dataset.}
\centering
\definecolor{grey}{RGB}{230,230,230}
\begin{tabular}{l|cccc|cc}
\hline
\multirow{2}{*}{Method} & \multicolumn{4}{c|}{\textbf{Image Quality (Novel View)}} & \multicolumn{2}{c}{\textbf{Render Quality}} \\ \cline{2-7}
                        & CLIP-Score ↑ & Sharp ↑ & Colorful ↑ & Quality ↑ & PSNR ↑ & SSIM ↑ \\ \hline
LucidDreamer~\cite{chung2023luciddreamer} & 0.656 & 0.961 & 0.603 & 0.704 & - & - \\
SceneDreamer360~\cite{li2024scenedreamer360} & 0.773 & 0.970 & \textbf{0.760} & 0.736 & 24.59 & 0.857 \\
\cellcolor{grey} Ours & \cellcolor{grey} \textbf{0.782} & \cellcolor{grey} \textbf{0.973} & \cellcolor{grey} 0.740 &  \cellcolor{grey} \textbf{0.747} & \cellcolor{grey} \textbf{25.74} & \cellcolor{grey} \textbf{0.885} \\ \hline
\end{tabular}
\label{table:stage1_quan}
\end{table*}



\subsection{3rd Stage: Temporal Alignment for 4D}
\label{sec:3rd}
\noindent\textbf{Motion Conditioned Video Generation}
At this stage, we aim to manipulate spatially aligned objects based on a user-defined 3D trajectory path of $(p_x,p_y, p_z)_{i=1...N}$. We begin by generating multi-view images projected from $P_{comp}$ by setting different azimuth and elevation angles for the camera views. For example, we define azimuth range as [0\degree, 90\degree, 180\degree, 270\degree] and elevation range as [0\degree, 30\degree], offering users unconstrained multi-views.
The 3D trajectory is then reprojected for each view, yielding total 8 pairs of multi-view images and corresponding 2D trajectories. We represent those pairs as ${(V_i, T_i)}_{i=1...8}$, where $V_i$ denotes each view image and $T_i$ its corresponding 2D trajectory. 
To generate video from image $V$ following a specified trajectory $T$, we opt DragAnything~\cite{wu2025draganything} as a motion-conditioned video model. We train the model on large-scale video dataset, VIPSeg~\cite{miao2022large}, using ControlNet~\cite{zhang2023adding} to condition trajectory following DIFT features. Specifically, the DIFT features of the video’s first frame are pooled based on each mask, allowing us to obtain an entity representation $\hat{E}$ $\in$ $\mathbb{R}^{H \times W \times C}$ along with a gaussian heatmap $h$.
They are mapped to the following frames according to the trajectory path. The objective to condition the motion into video diffusion model can be simplified to:
\begin{equation}
    \label{eqn:denoise}
    \mathcal{L}_{\theta} = \sum_{i=1}^{L} \left\| \epsilon - \epsilon_{\theta} \left( z, \mathcal{E}_{\theta}(\hat{\mathbf{E}}_i), \mathcal{E}_{\theta}(h_i) \right) \right\|_{2}^{2}
\end{equation}
where latent feature of first frame ($z$) and encoded trajectory features ($\mathcal{E}_{\theta}(\hat{\mathbf{E}}_i), \mathcal{E}_{\theta}(h_i)$) with encoder $\mathcal{E}$ are added to the denoised features in diffusion model. However, as shown in the first row of the second box in \cref{fig:stage2_qual}, we observe motion hallucination in a local part of the moving object. To address this, we augment the current DIFT features from the instance mask with a part-level DIFT feature derived from the part mask of each instance. For VIPSeg, consistent with the previous stage, we use DINOv2~\cite{oquab2023dinov2} and apply k-means clustering within the instance mask to obtain the part mask. This enables us to extract part features, which are then concatenated with the global feature and their respective trajectories, represented as $\hat{E^{\text{part}}_i}$ and $h^{\text{part}}_i$. Consequently, \cref{eqn:denoise} can be modified as follows:
\begin{equation}
    \scriptsize
    \label{eqn:part_denoise}
    \mathcal{L}_{\theta} = \sum_{i=1}^{L} \left\| \epsilon - \epsilon_{\theta} \left( z, \mathcal{E}_{\theta}(\hat{\mathbf{E}}_i), \mathcal{E}_{\theta}(h_i), \mathcal{E}_{\theta}(\hat{\mathbf{E^{\text{part}}}}_i), \mathcal{E}_{\theta}(h^{\text{part}}_i) \right) \right\|_{2}^{2}
\end{equation}
\begin{table}[t!]
\caption{Ablation Studies of Object Scene Composition. There is trade-off between geometric quality and image quality depending on usage of SDS loss and joint learning of background.}    \centering
    \resizebox{1.0\linewidth}{!}{
    \setlength{\tabcolsep}{3pt} 
    \begin{tabular}{l c c c c | c c | c }
\thickline
     Method & SDS & Normal. C & w/o BG & Rasterize & CLIP-Score ↑ & Sharp ↑ & Align ↓  \\ 
    \hline
        (1) &\redxmark & \greencheckmark & \redxmark & \greencheckmark & 0.643  & 0.971 & \textbf{0.134}  \\  
        (2) &\greencheckmark & \greencheckmark & \redxmark & \redxmark & 0.661 & 0.973 & 0.265  \\
        (3) &\greencheckmark & \redxmark & \greencheckmark & \greencheckmark & 0.753 & \textbf{0.977} & 0.545  \\
        (4) &\greencheckmark & \greencheckmark & \redxmark & \greencheckmark & \textbf{0.775} & 0.976 &  0.529 \\
        (5) &\greencheckmark & \greencheckmark & \greencheckmark & \greencheckmark & \textbf{0.775}  & 0.976  & 0.528   \\
        \thickline
    \end{tabular}}
    \label{tab:ablation_study}
\end{table}



This modified objective serves as the overall training goal for our part-augmented, motion-conditioned video generation. As the model aims to align motion by considering both local and global features of instances, we refer to it as \textbf{L}ocal-\textbf{G}lobal DragAnything.
Given the pretrained LG-DragAnything and desired trajectory from users, we generate both global- and part-level trajectory features from ${(V_i, T_i, F_i)}$, enabling the synthesis of eight motion-conditioned videos with a single model. Here, $F_i$ represents the DIFT feature map, partitioned at the part and instance levels. From the optimized object–scene composited 2D-GS, we rasterize ${(V_i, T_i, F_i)}$ for each view and leverage these priors to produce spatially and temporally coherent 4D videos with LG-DragAnything.







\section{Experiments}
In this section, we evaluate Drag4D across three key tasks: Text-to-3D Generation (\cref{sec:t23d}), Object-Scene Composition (\cref{sec:compose}), and Motion-Conditioned Video Generation (\cref{sec:vcvideo}), each aligning with one of Drag4D’s distinct stages. We provide the details of baselines used for those three tasks, main quantitative and qualitative results, and ablation studies.

\noindent\textbf{Datasets} Due to the absence of publicly available datasets for validating 4D environments with motion guidance, we created our own dataset, named  \textit{Drag4D-30}. The dataset comprises 30 complex and extended text prompts designed to synthesize corresponding 3D scenes, demonstrating the effectiveness of Drag4D compared to other baselines in 1st stage. Additionally, it includes 4 object-centric images, each containing a target instance. These images are used for evaluating the 2nd stage and 3rd stage, where they are paired with the 30 text prompts to assess the spatial and motion alignment of the object within the 30 3D scenes. We will provide the details of \textit{Drag4D-30} in supplementary material.

\subsection{Text-to-3D Generation}
\label{sec:t23d}
\noindent\textbf{Baselines} As 1st stage of our Drag4D aims to reconstruct 3D scene from text prompt, we compare our approach with two recent methods. 1) LucidDreamer~\cite{chung2023luciddreamer}, employs a technique where outpainted RGBD images are mapped onto a point cloud, which is then used to guide the optimization of 3DGS by projecting various images derived from this point cloud. However, as LucidDreamer lacks the capability to directly produce 3D scenes from textual prompts, we address this limitation by leveraging a diffusion model to create conditional images, enabling the generation of 3D scenes based on text input.
2) SceneDreamer360\degree~\cite{li2024scenedreamer360}, utilizes a text-driven panoramic image generation model, fine-tuned with a three-stage enhancement process, to produce high-resolution panoramas. These panoramas are integrated into 3D space using 3D-GS, ensuring multi-view consistency.

\begin{table}[t!]
\caption{Performance comparison of motion-conditioned video generation on VIPSeg validation set. Our proposed LG-DragAnything with part augmentation surpasses the baseline across both image-based metrics (FID, PSNR, SSIM) and video-based metrics (FVD and ObjMC). Notably, higher values indicate better performance for PSNR and SSIM, while lower values are preferable for FID, FVD, and ObjMC. Results marked with $\ast$ indicates that we reproduce better score from baseline, DragAnything~\cite{wu2025draganything}.}
\centering
\definecolor{grey}{RGB}{230,230,230}
\scalebox{0.7}{
\setlength{\tabcolsep}{4pt} 
\begin{tabular}{l|ccccc}
\hline
Method            & FID ↓ & FVD ↓   & PSNR ↑ & SSIM ↑ & ObjMC ↓ \\
\hline
$\ast$DragAnything~\cite{wu2025draganything}    & 34.45 & 288.68  & 18.41  & 0.57   & 19.9\\
\cellcolor{grey}LG-DragAnything       & \cellcolor{grey}32.79 & \cellcolor{grey}272.02  & \cellcolor{grey}19.02  & \cellcolor{grey}0.59   & \cellcolor{grey}17.6    \\
\hline
\end{tabular}}
\label{table:stage3_quan}
\end{table}

\begin{figure}[t!]
\centering
\includegraphics[width=1.01\linewidth]{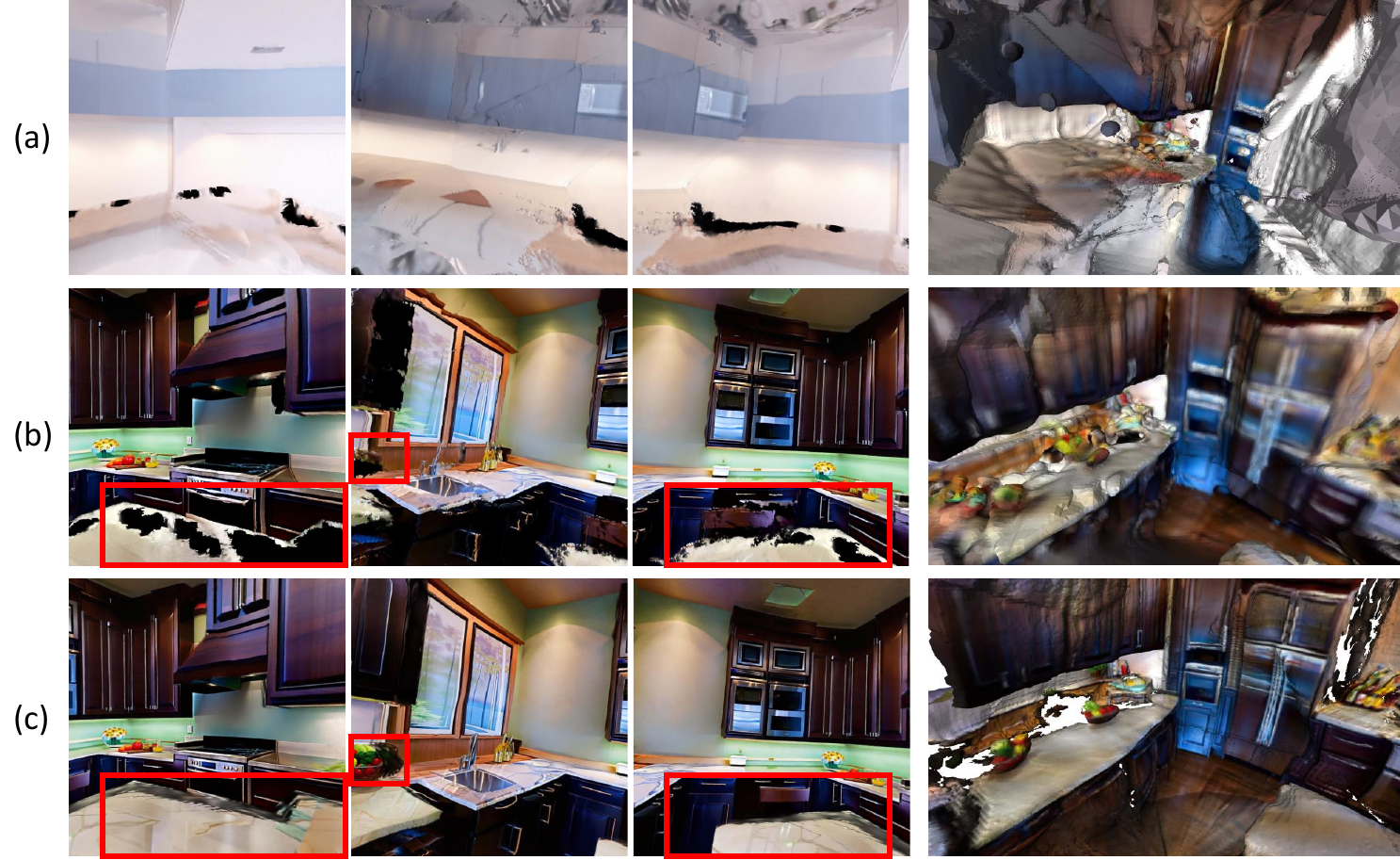}
\caption{Qualitative Results on the 1st Stage with Drag4D dataset. We show rendered color images in novel viewpoint and mesh reconstructed from (a) LucidDreamer~\cite{chung2023luciddreamer} (b) SceneDreamer360~\cite{li2024scenedreamer360}, and (c) ours. Our method can effectively handle unseen viewpoints due to our adaptive inpainting strategy. It is best viewed in color and high resolution; please zoom in .}
\label{fig:stage1_qual}
\end{figure}

\noindent\textbf{Main Results}
To assess the fidelity of the generated 3D scenes to the text prompts, we calculate the CLIP-Score~\cite{hessel2021clipscore}. Additionally, we use CLIP-IQA~\cite{wang2023exploring} to evaluate visual sharpness, colorfulness, and quality. Both metrics leverage the pre-trained CLIP-B/32 model~\cite{radford2021learning}. Furthermore, PSNR and SSIM are employed to measure rendering quality. As presented in \cref{table:stage1_quan}, Drag4D surpasses the baselines in both image quality and rendering quality. The qualitative results in \cref{fig:stage1_qual}
 demonstrate that Drag4D produces visually complete and less distorted 3D scenes, attributed to our joint training with base images and inpainting-augmented views.

\begin{table}[t!]
\caption{Comparison between the DragAnything~\cite{wu2025draganything} and our LG-DragAnything on \textit{Drag4D-30} dataset.}
\centering
\definecolor{grey}{RGB}{230,230,230}
\scalebox{0.75}{
\begin{tabular}{l|ccccc}
\hline
Method            & CLIP-Score ↑ & Quality ↑ & Colorful ↑ \\
\hline
$\ast$DragAnything~\cite{wu2025draganything}  & 0.805 & 0.48 &  0.71 \\
\cellcolor{grey}LG-DragAnything       &  \cellcolor{grey}0.814 & \cellcolor{grey}0.51  & \cellcolor{grey} 0.75 \\
\hline
\end{tabular}}
\label{table:comparison3}
\end{table}
\subsection{Object-Scene Composition}
\label{sec:compose}
\begin{figure}[t!]
\centering
\includegraphics[width=1.0\linewidth]{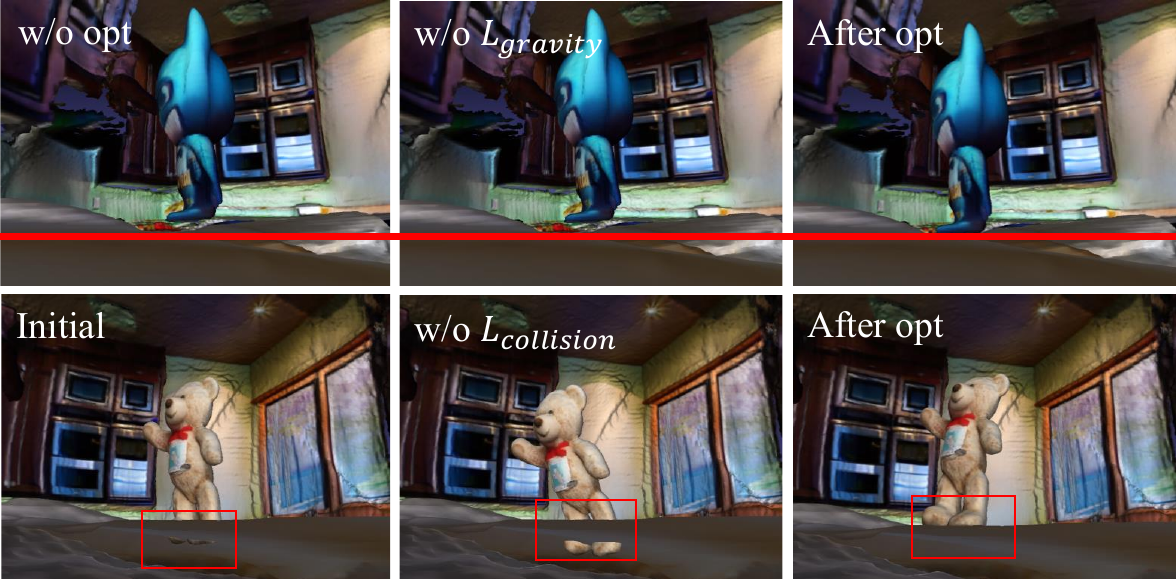}
\caption{Ablation study of physics-aware position learning used in object-scene composition}
\label{fig:composition1}
\end{figure}

\noindent\textbf{Baselines and Main Results} Since there is previous baseline in object-scene composition, we construct our self-baseline as summarized in \cref{tab:ablation_study}. We can observe that using SDS loss and normal consistency loss help to increase both CLIP-Score and Sharpness. Additionally, according to \cref{fig:composition1}, we can easily find out that using both collision loss and gravity loss help to position the object accurately. 


\subsection{Motion-Conditioned Video Generation}
\label{sec:vcvideo}

\noindent\textbf{Baselines} We chose DragAnything~\cite{wu2025draganything} as a representative baseline to evaluate motion-conditioned video generation. It proposes a framework for controllable video generation that uses entity representationhs for motion control of any object. It enables trajectory-based interaction, removing the need for additional guidance signals like masks or depth maps. Our proposed Drag4D introduces part augmentation strategy on top of DragAnything. Both methods are trained with VIPSeg~\cite{miao2022large} training datasets.


\begin{figure}[t!]
\centering
\includegraphics[width=1.05\linewidth]{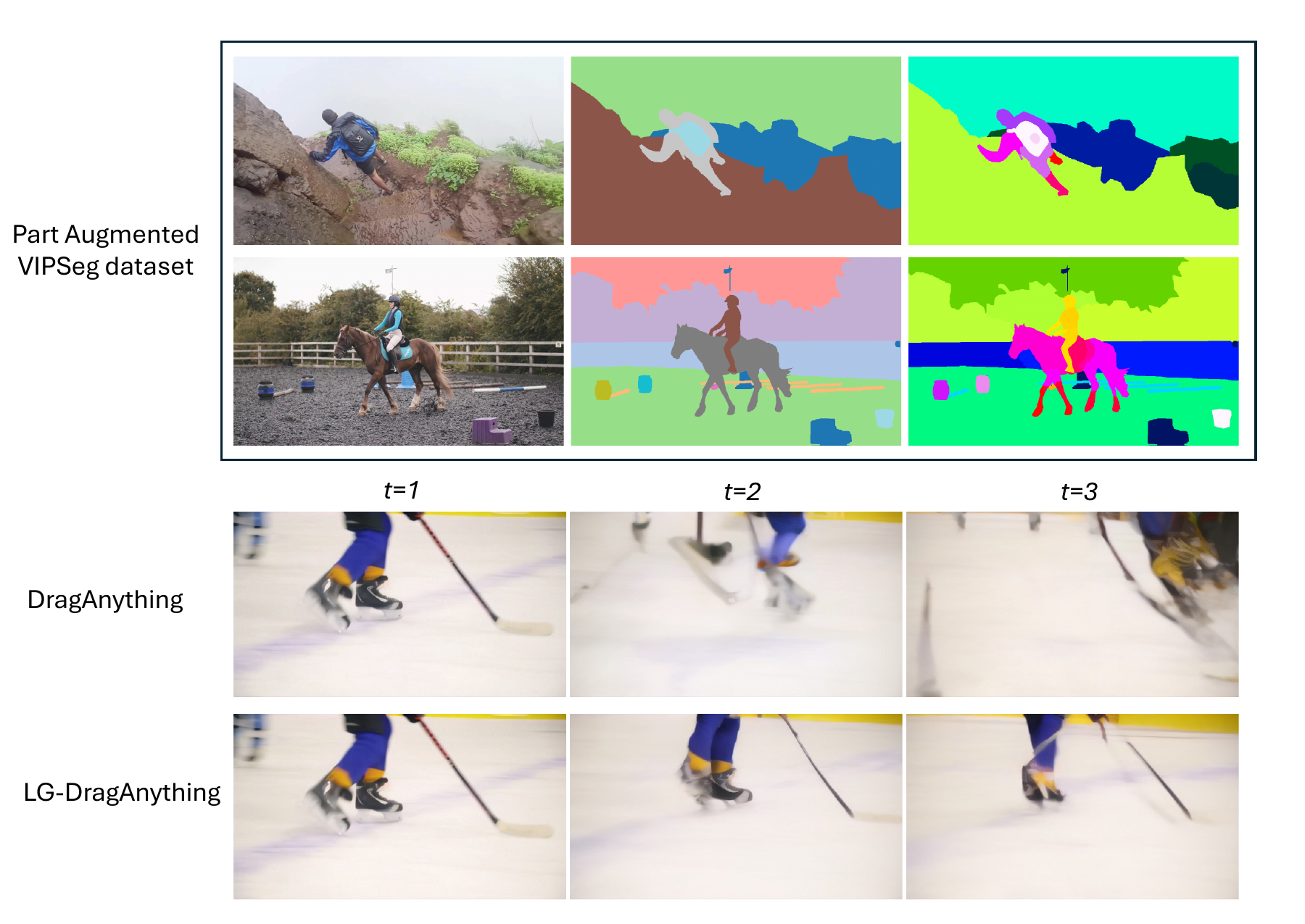}
\caption{Qualitative Results on the 3rd Stage with VIPSeg dataset~\cite{miao2022large}. We adapt the VIPSeg dataset by annotating it with part segmentation, achieved through feature clustering from DINOv2~\cite{oquab2023dinov2}. This modified dataset is then used to train a part-augmented, motion-conditioned video model as described in \cref{eqn:part_denoise}. Our results show that LG-DragAnything with part augmentation effectively reduces motion hallucination by accounting for motion at both global and part levels.
Best viewed in color and high resolution; please zoom in for finer details.}
\label{fig:stage2_qual}
\end{figure}

\begin{figure*}[t!]
\centering
\includegraphics[width=0.89\linewidth]{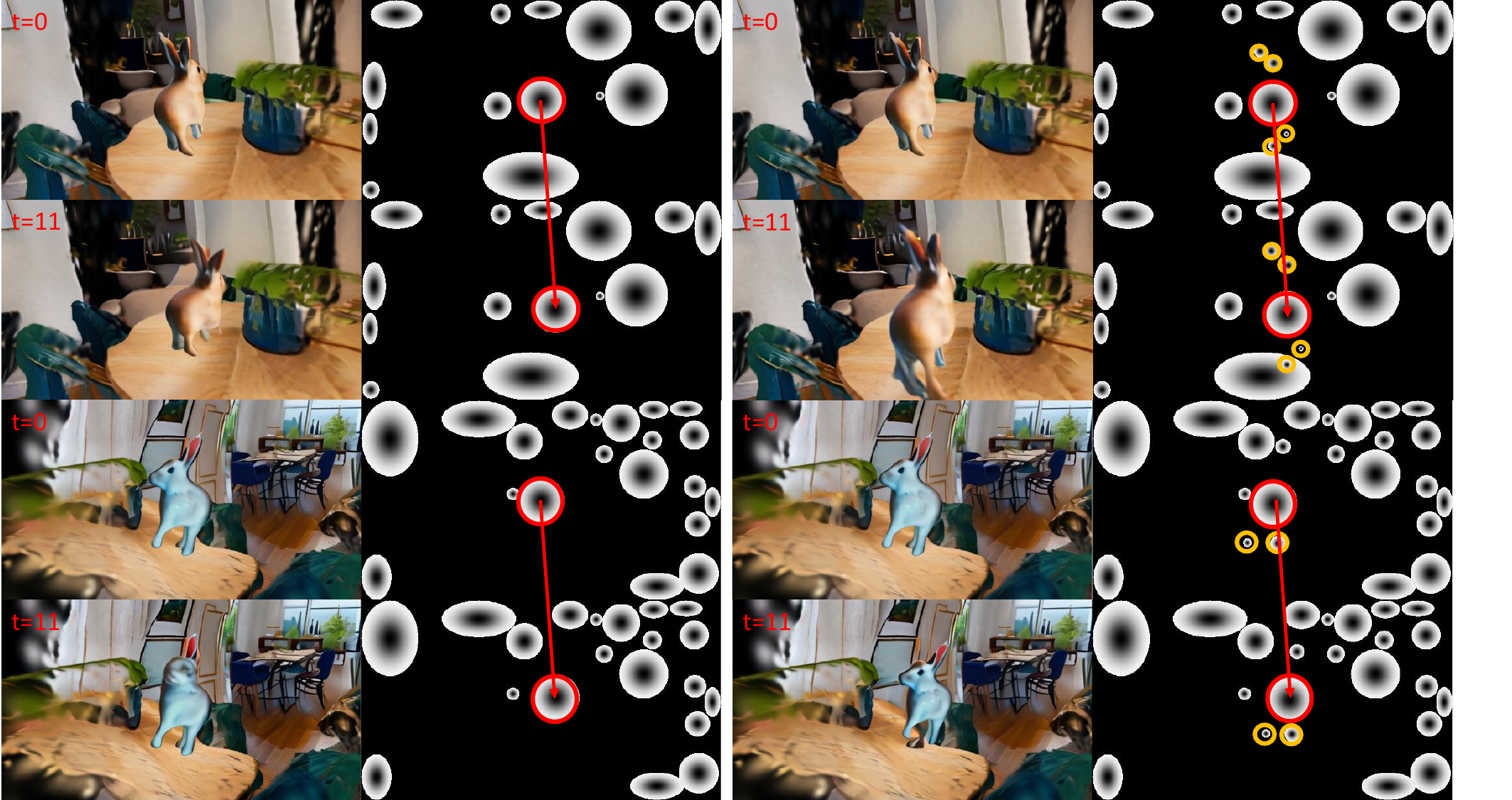}
\caption{Qualitative Results on the 3rd Stage, which is our multi-view generated video result. Left is from DragAnything and right is from our LG-DragAnything. Part guidance leads to clear and intended results.} 
\label{fig:result1}
\end{figure*}

\begin{figure*}[t!]
\centering
\includegraphics[width=0.89\linewidth]{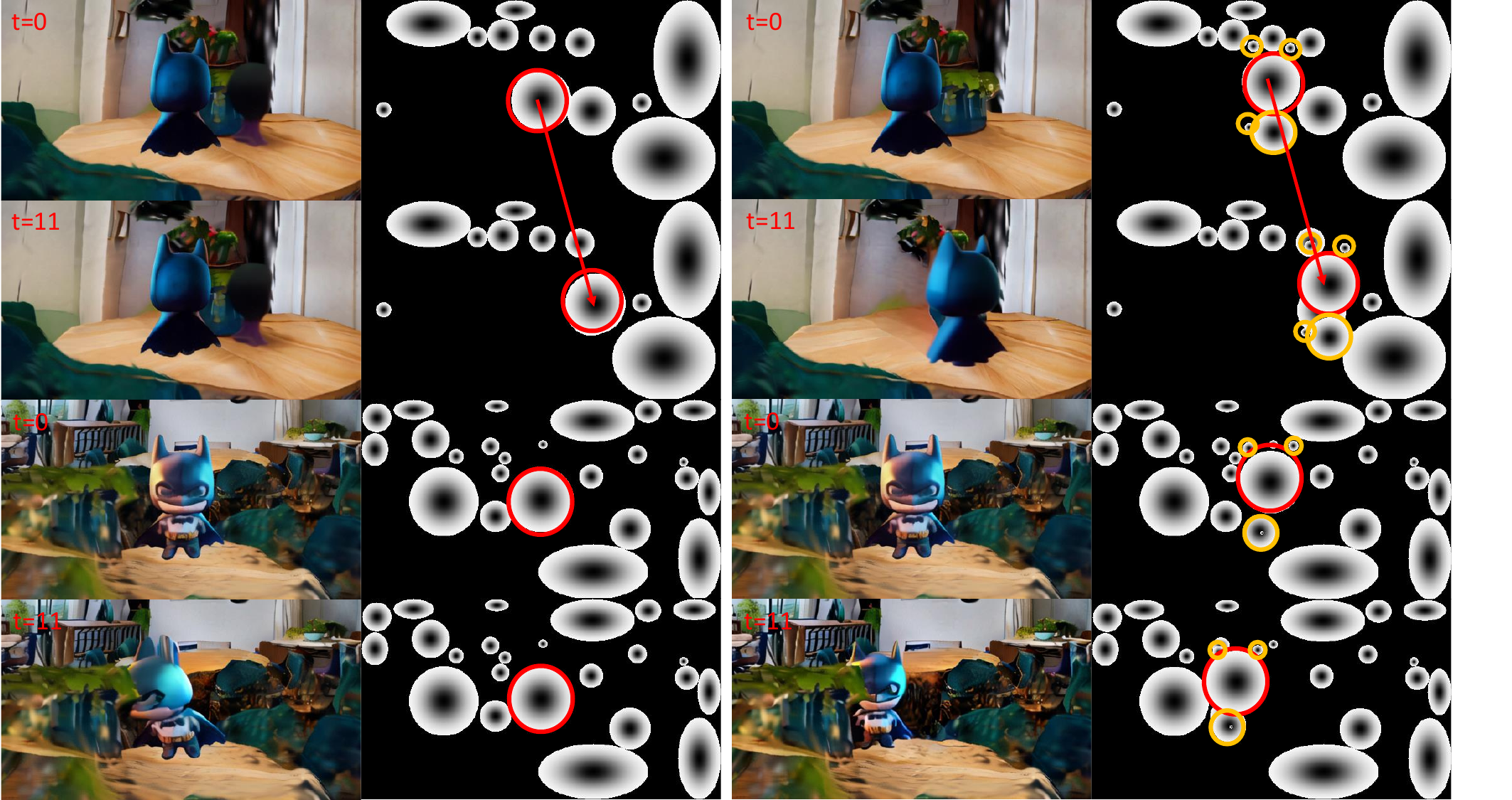}
\caption{ Qualitative Results on the 3rd Stage, which is our multi-view generated video result. Left is from DragAnything and right is from our LG-DragAnything. Part guidance leads to clear and intended results.} 
\label{fig:result2}
\end{figure*}

\noindent\textbf{Main Results} We evaluate the motion alignment of our part augmentation method compared to DragAnything in two scenarios: 1) motion alignment in 2D videos: Using the VIPSeg validation set, we assess performance based on image metrics (e.g., FID, PSNR, and SSIM) and video metrics (e.g., FVD and ObjMC). As shown in \cref{table:stage3_quan}, part augmentation demonstrates improved performance across these metrics. 2) motion alignment in 4D videos: We measure CLIP-Score to evaluate text fidelity and utilize CLIP-IQA metrics, including quality and colorfulness, to assess the quality of multi-view videos on \textit{Drag4D-30} dataset. We summarize the quantitative result in \cref{table:comparison3}. \cref{fig:stage2_qual} shows visual impact of LG-DragAnything in the first scenario, while \cref{fig:result1} and \cref{fig:result2} correspond to the second scenario.
The results from these two scenarios demonstrate that our proposed part-augmentation in motion-conditioned video generation effectively reduces local hallucinations while improving fidelity to the text prompt.






\section{Conclusion}
We introduce \textbf{Drag4D}, a comprehensive interactive pipeline designed to align user-defined 3D object motion with text-driven 3D background scene generation. In the first stage, Drag4D generates a high-fidelity 3D scene by optimizing 2D Gaussian representations on panoramic images and their augmented views, surpassing previous state-of-the-art models in 3D scene generation. In the second stage, Drag4D extracts the target instance from a user-provided reference image, transforming it into a full 360° object that is spatially aligned with the generated 3D scene using our proposed 3D Copy-and-Paste method. The final stage further enhances user experience, allowing temporal manipulation of the 3D object within the 3D scene using a part-augmented motion-conditioned video generator and 4D Gaussian representations. We anticipate significant societal benefits from Drag4D, as it performs robustly across diverse user prompts, offering potential applications in fields such as entertainment, video synthesis and AR/VR.

{
    \small
    \bibliographystyle{ieeenat_fullname}
    \bibliography{main}
}

\clearpage
\appendix
\onecolumn
\section*{\Large Supplementary Material}

\section{Related Works}
\label{sec:supple_related}
\noindent\textbf{3D Scene Generation.}\quad
3D Scene generation has been actively studied due to the rapid development of image generation. Early studies~\cite{schwarz2020graf,chan2021pi,niemeyer2021giraffe,nguyen2019hologan,nguyen2020blockgan} utilize Generative Adversarial Networks (GANs) and implicit neural networks to represent 3D objects with texture. However, these methods have limited ability to generate diverse categories of objects and scenes due to GAN's inherent difficulty of learning and limited 3D representation. Advanced recent studies~\cite{hollein2023text2room,poole2022dreamfusion,chung2023luciddreamer,zhang20243ditscene} generate large 3D scenes from either text prompts or a single image from the user by incorporating advanced diffusion-based image generation techniques~\cite{song2020denoising,rombach2022high}. These methods use Neural Radiance Fields (NeRFs)~\cite{hollein2023text2room,poole2022dreamfusion} to represent 3D scenes or 3D Gaussian Splatting~\cite{chung2023luciddreamer,zhang20243ditscene,shriram2024realmdreamer} for creating high-fidelity results and efficient generation. To generate 3D scene with a large field of view, these methods use diffusion prior to progressively outpaint the unseen part of the scene and integrate it to get the full 3D scene. The combined 3D scene from these outpainting steps often suffers from multi-view inconsistencies, as diffusion priors struggle to maintain coherence across different camera viewpoints. To alleviate this limitation, some recent methods~\cite{li2024scenedreamer360,ma2024fastscene,zhou2024holodreamer,zhou2025dreamscene360} generate panoramic images from text prompt and learn to reconstruct 3D scene using 3D Gaussian Splatting. Compared to these recent methods, our method utilizes 2D Surfel Gaussian, takes advantage of high-quality geometry reconstruction, and also inpaints unseen parts of augmented viewpoints in the training stage so that our method shows robustness under unseen novel viewpoints as well.

\noindent\textbf{Object Scene Composition.}\quad
Given the desired 3D layout positions of multiple objects, there have been several attempts to generate these objects together within a cohesive scene. CG3D~\cite{vilesov2023cg3d} enables physically realistic composition and generation of multiple objects by using physics-inspired losses. GraphDreamer~\cite{gao2024graphdreamer} employs scene graphs to represent relationships between multiple objects, ensuring the generated scene follows these relational constraints. More recent studies~\cite{li2024dreamscene,zhang2024towards} have explored compositing scenes and objects derived from text-to-3D models. However, the textures produced by these methods often lack realism due to their heavy dependence on diffusion models or CLIP priors. Moreover, we employ two-stage optimization inspired by~\cite{zhou2024layout}, first optimizing the object's position using physics prior and jointly training positioned objects with a pre-trained scene for the natural composition.

\noindent\textbf{Controllable Video Generation.}\quad
With advancements in diffusion models significantly improving video generation performance, there has been a growing interest in controllable video generation. While numerous existing studies focus on generation conditioned on text, images, depth, or skeletal data, our work is specifically aligned with video generation conditioned by either camera movement or motion trajectories. DragNUWA~\cite{yin2023dragnuwa} proposes a multi-scale trajectory encoding approach that integrates trajectory conditioning into a video diffusion model, along with adaptive training to effectively learn from dense optical flow to more intuitive, user-friendly motion. DragAnything~\cite{wu2025draganything} introduces an entity representation to enable instance-level motion guidance, effectively mitigating distortions and undesired deformations often associated with point-based trajectories. Building on these approaches, CameraCtrl~\cite{he2024cameractrl} use Plücker coordinates to precisely control camera trajectories, while MotionCtrl~\cite{wang2024motionctrl} goes further by decomposed control of both camera and object trajectories. Our work builds upon DragAnything~\cite{wu2025draganything}, extending it to allow part-based instance control. We observed that controlling only at the instance level is insufficient for articulated objects and often leads to hallucinations in the generated result.

\section{Implementation Details}
\label{sec:supple_implementation}
In this section, we aim to elaborate on the details of Stage 1 and Stage 2, which were not fully covered in the manuscript.

\noindent\textbf{Stage 1. 3D Scene Generation.}\quad
\begin{figure*}[t]
\centering
\includegraphics[width=1.0\linewidth]{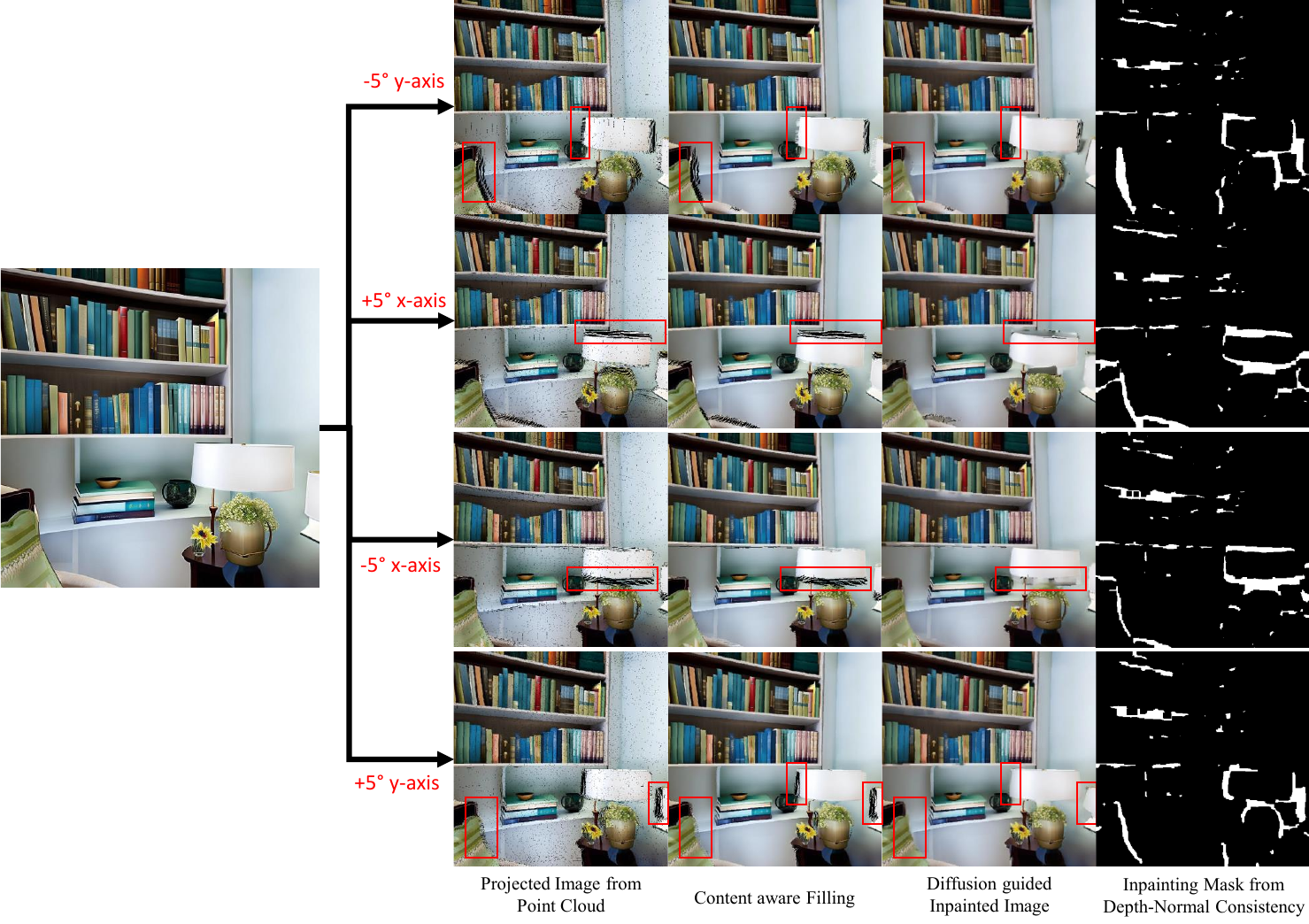}
\caption{\textbf{Viewpoint Augmentation in Stage 1}: We describe our viewpoint augmentation strategy adapted in stage1 to reconstruct 3D scene from given panorama image. First, the point cloud is projected onto four augmented viewpoints, where the consistency of the projected depth and normals is evaluated. A geometric uncertainty map derived from this evaluation guides the inpainting process, addressing unseen and distorted pixels.}
\label{fig:supple_inpaint}
\end{figure*}
Given the panoramic image generated from the text, we follow our baselines~\cite{zhou2024holodreamer,li2024scenedreamer360} to augment viewpoints additional to the sphere projected images from the panorama. This is because the projected images only offer very limited camera viewpoints, leading the reconstructed scene to be overfitted to that viewpoint and significantly distorted when viewed from novel perspectives. Unlike previous works~\cite{zhou2024holodreamer,li2024scenedreamer360}, our approach directly generates images at augmented viewpoints through a combination of view projection and inpainting steps depicted in~\cref{fig:supple_inpaint}. Specifically, we begin by projecting the globally aligned point cloud, obtained using methods from~\cite{rey2022360monodepth,yin2023metric3d} and following the projection process in~\cite{zhou2024holodreamer}, to create a projected image with holes. Next, we use content-aware filling~\cite{bertalmio2001navier} to fill these holes. To evaluate consistency, we calculate the cosine similarity between the projected depth and normals derived from the point cloud. Finally, we define an uncertainty mask for regions with similarity below 0.75 and in-paint pixels within these uncertain areas.

\begin{figure}[t]
\centering
\includegraphics[width=1.0\linewidth]{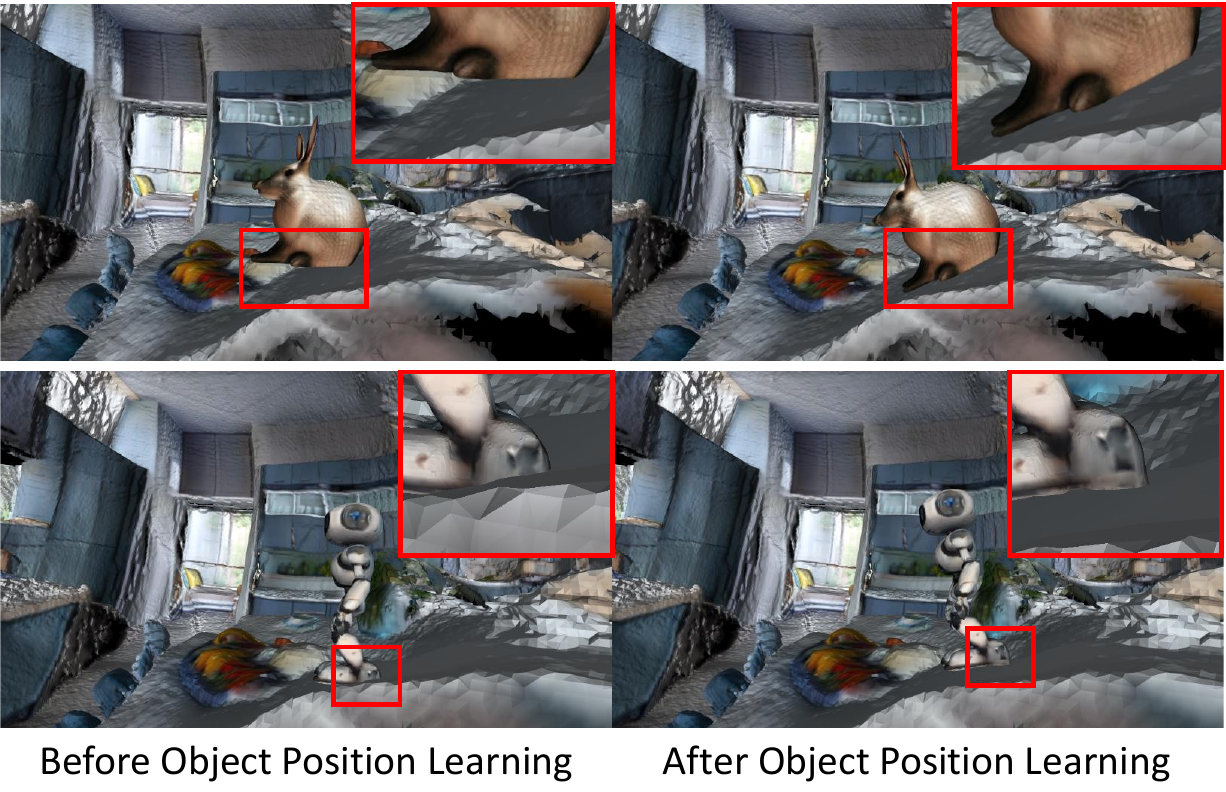}
\caption{\textbf{Object Position Learning in Stage 2}: In stage 2, object position learning is crucial in aligning the object's position to seamlessly fit within the given scene. We demonstrate the visual difference between including this learning step and not.}
\label{fig:supple_comp}
\end{figure}

\noindent\textbf{Stage 2. Object Scene Composition.}\quad
We show our physics-aware object-scene composition framework's effectiveness in~\cref{fig:supple_comp}.

\noindent\textbf{Hyperparameters.}\quad
When we train 2D-GS in stage 1, we use the same parameters as SceneDreamer360~\cite{li2024scenedreamer360} and train 4000 iterations in total. We use the learning rate of object position learning as 0.001.


\section{Drag4D-30 Dataset}
\label{sec:supple_drag4d}
\begin{figure*}[t]
\centering
\includegraphics[width=1.0\linewidth]{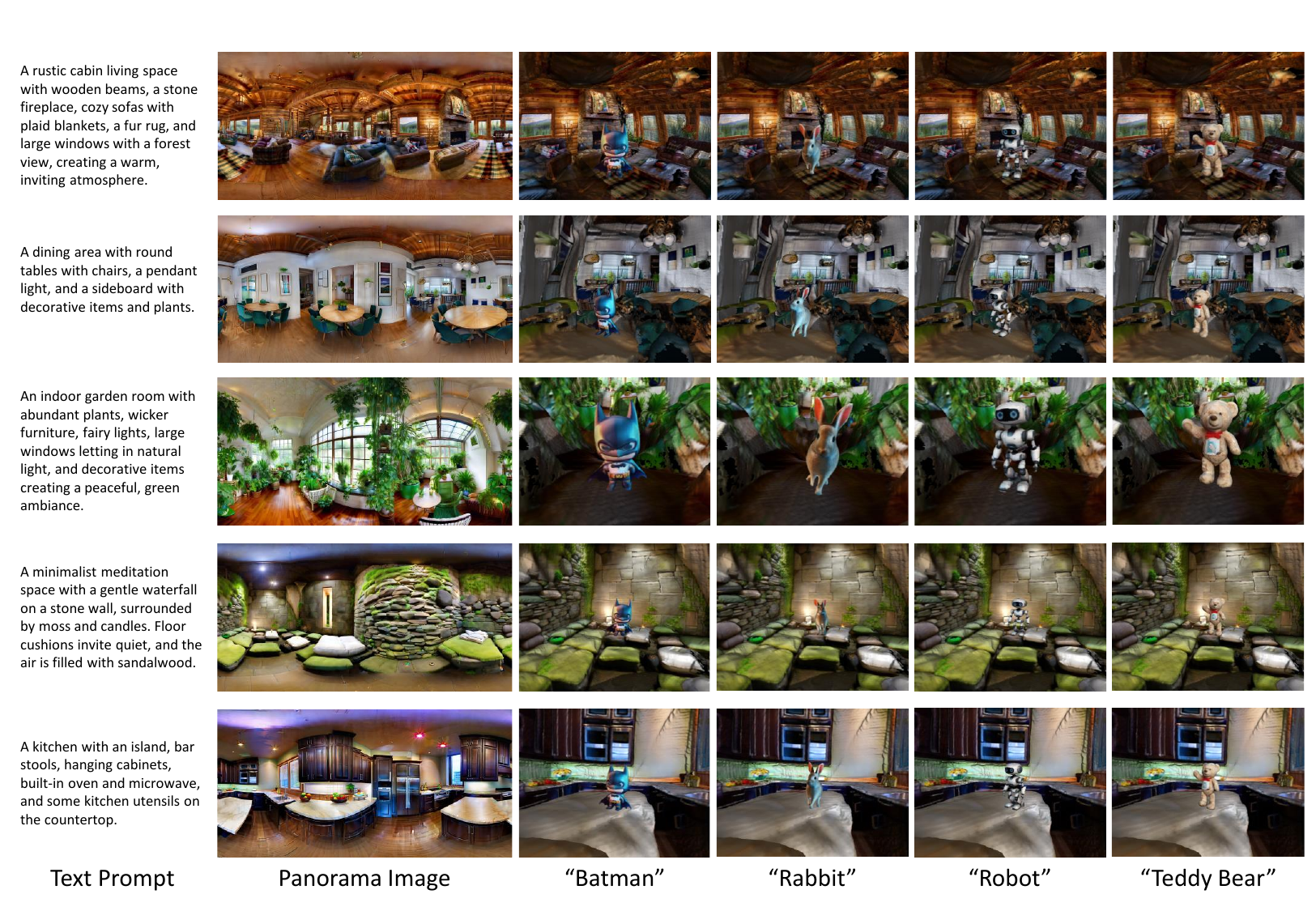}
\caption{\textbf{Drag4D-30 Dataset}: We visualize some of our aligned 3D scenes with objects in 3D. Our scene is generated and reconstructed from the following text prompts. Thanks to our physically plausible object position learning step, our object is well-composited with the reconstructed scene. \textbf{Please take a closer look to observe the finer details.}}
\label{fig:supple_dataset}
\end{figure*}

Our Drag4D-30 Dataset features 30 distinct 3D scenes, each containing four different objects: "teddy bear," "batman," "rabbit," and "robot." To obtain these object assets, we generate 3D textured meshes from single images using InstantMesh~\cite{xu2024instantmesh} and DALLE3~\cite{betker2023improving}. For constructing the 3D scenes, we first generate 30 different panoramic images and corresponding text prompts using ChatGPT-4o~\cite{openai2024gpt4o} and PanFusion~\cite{panfusion2024}. Subsequently, we reconstruct 3D scene using 2D-GS~\cite{huang20242d}, described in stage 1 of the manuscript, to produce complete 3D scenes from these panoramic images. Finally, we visualize some samples of our Drag4D-30 dataset including text prompt, panorama image, and a mesh of the generated scene with aligned 4 objects (result after stage 1 training is finished), in~\cref{fig:supple_dataset}.

\section{Additional Results}
\label{sec:supple_results}
\begin{figure*}
\centering
\includegraphics[width=1.0\linewidth]{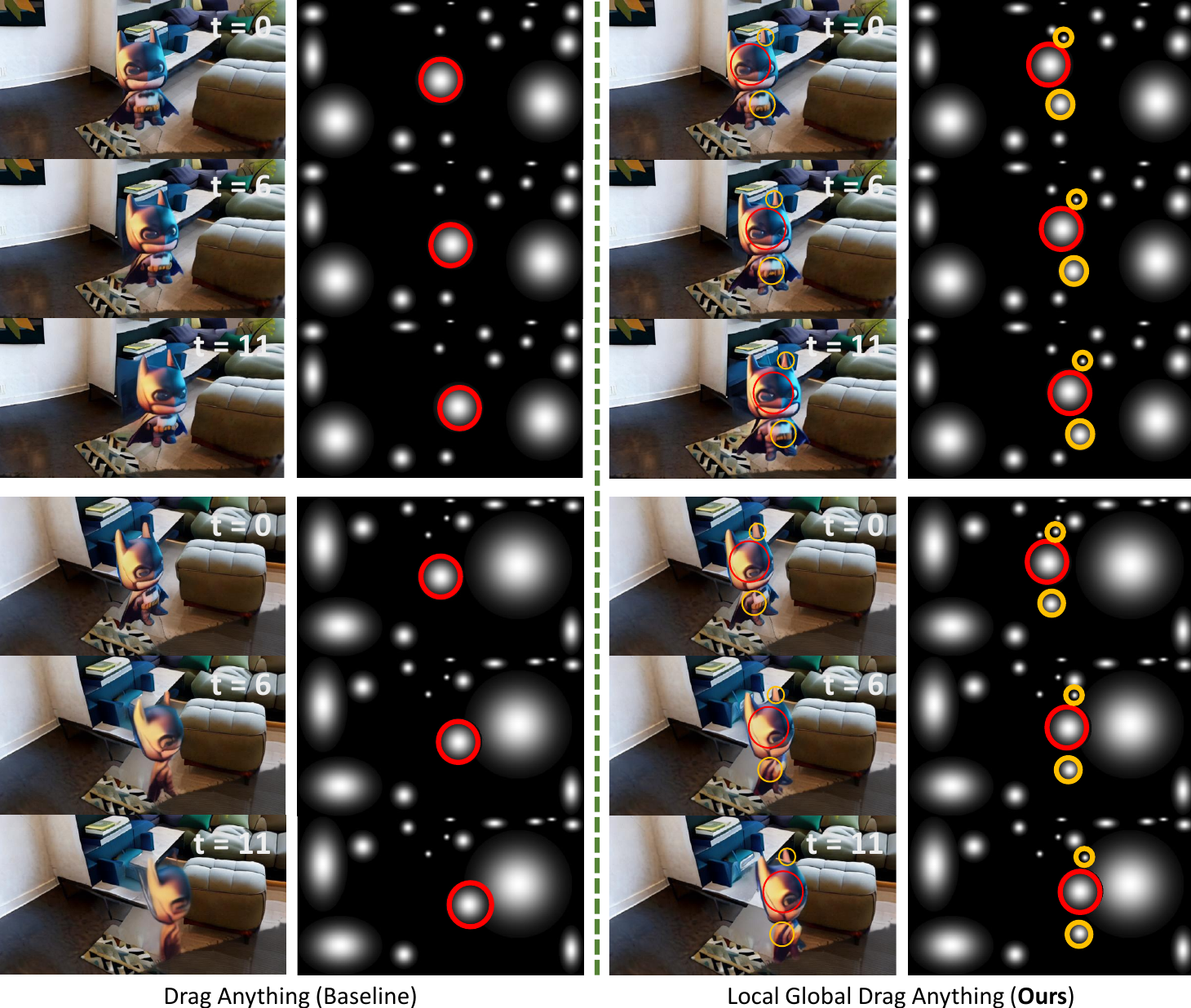}
\caption{Further qualitative comparisons between our baseline and LG-DragAnything are provided. We denote \textcolor{orange}{\textbf{part motion}} and \textcolor{red}{\textbf{global motion}} for the reader's understanding. Our method effectively guides both global and part movements, ensuring strict compliance with the given trajectory without hallucination and distortion.}
\label{fig:supple_comp1}
\end{figure*}

\begin{figure*}
\centering
\includegraphics[width=1.0\linewidth]{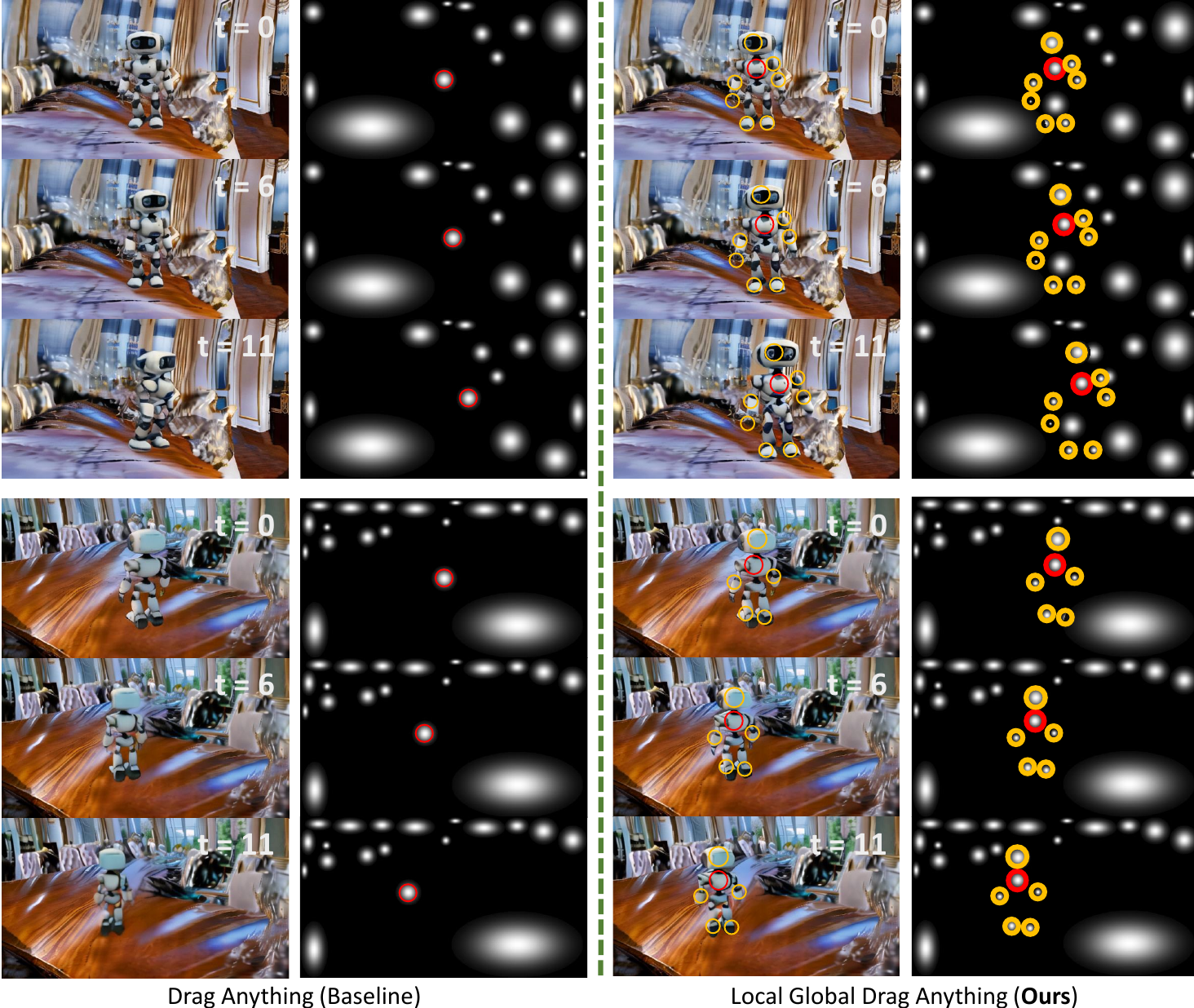}
\caption{Further qualitative comparisons between our baseline and LG-DragAnything are provided. We denote \textcolor{orange}{\textbf{part motion}} and \textcolor{red}{\textbf{global motion}} for the reader's understanding. Our method effectively guides both global and part movements, ensuring strict compliance with the given trajectory without hallucination and distortion.}
\label{fig:supple_comp2}
\end{figure*}

\begin{figure*}
\centering
\includegraphics[width=1.0\linewidth]{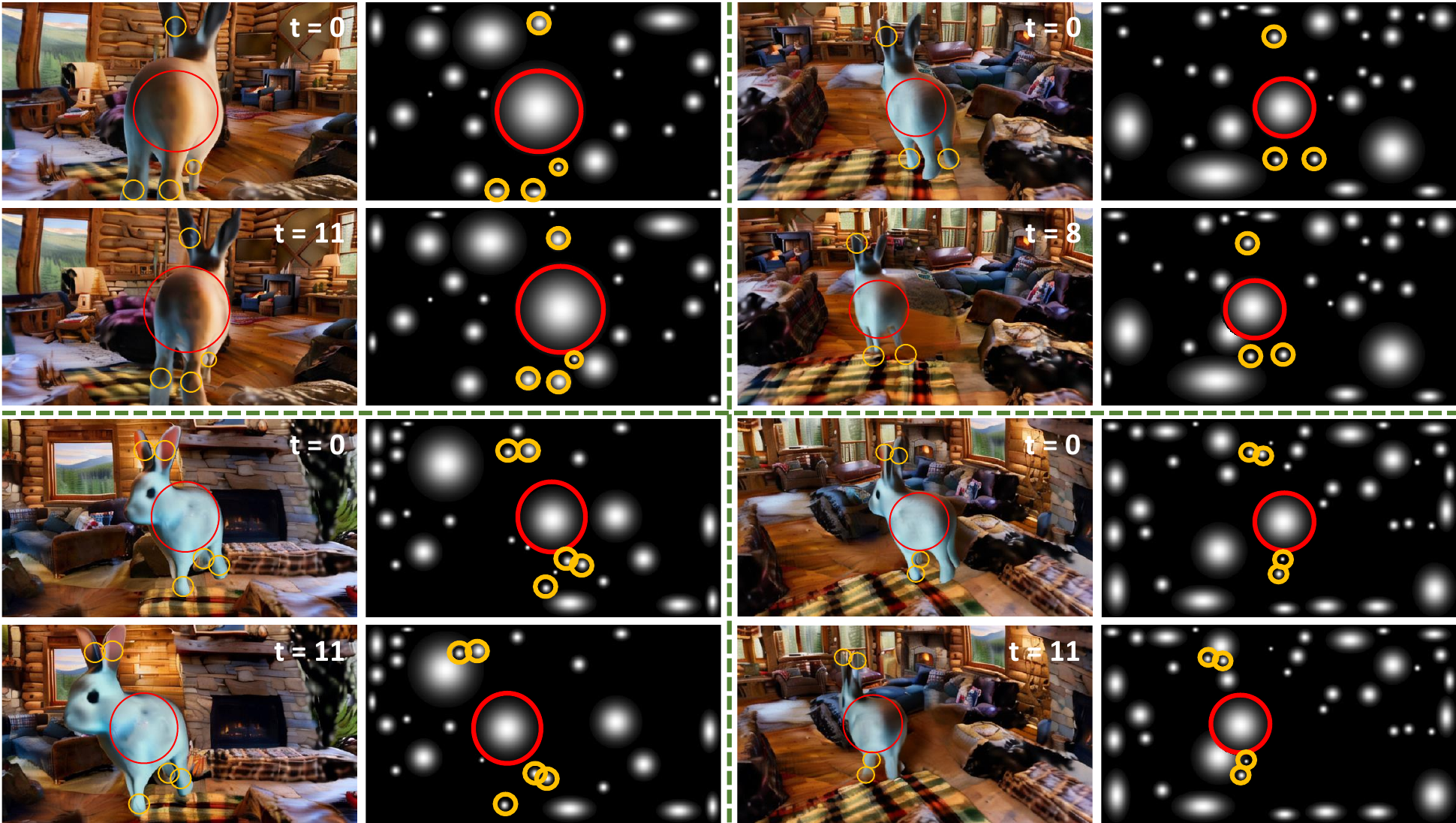}
\caption{A multi-view generated sequence of a moving rabbit moving in the 'cabin space' scene.}
\label{fig:supple_qual1}
\end{figure*}

\begin{figure*}
\centering
\includegraphics[width=1.0\linewidth]{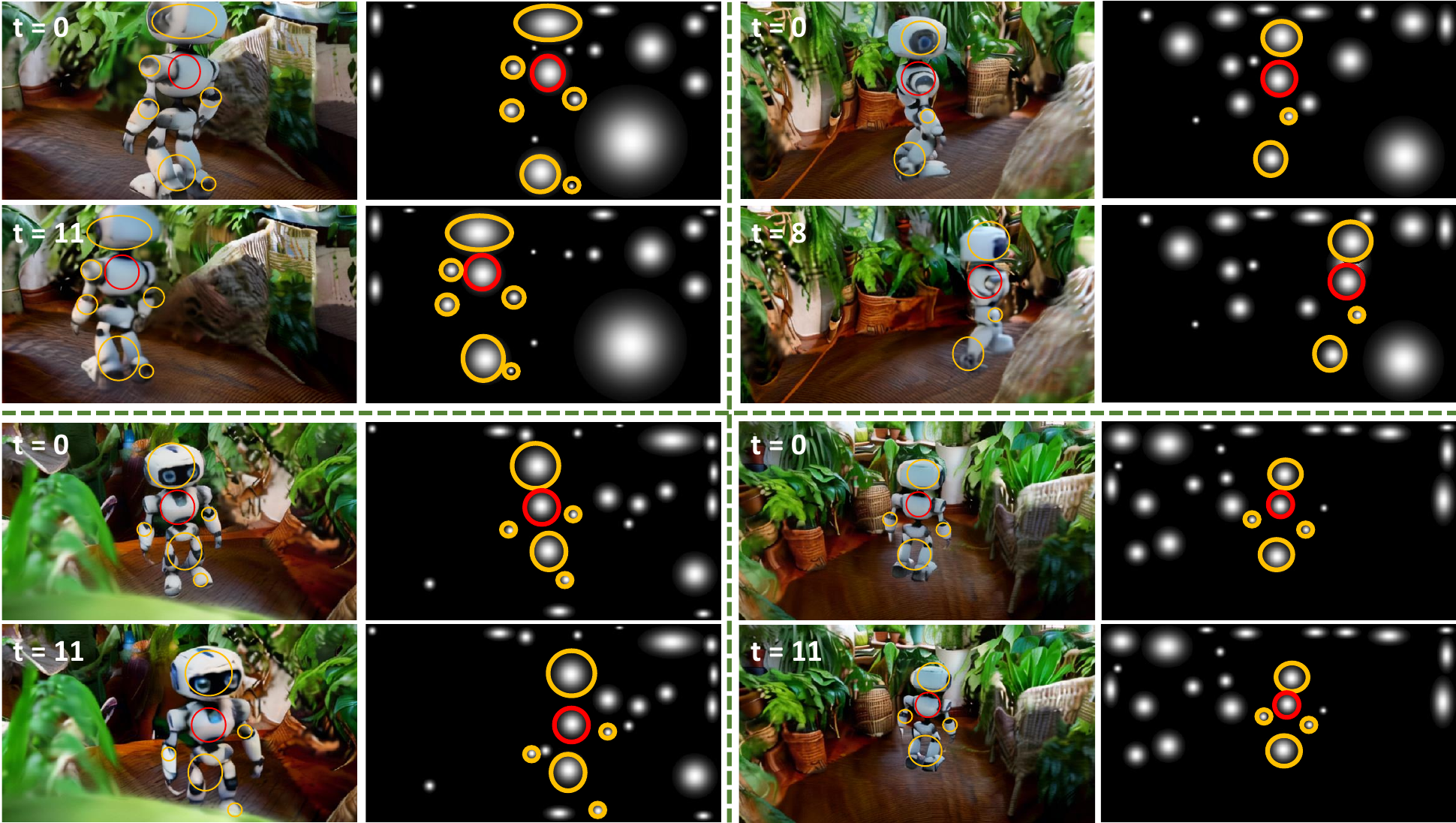}
\caption{A multi-view generated sequence of a moving robot moving in the 'garden room' scene.}
\label{fig:supple_qual2}
\end{figure*}

\begin{figure*}
\centering
\includegraphics[width=1.0\linewidth]{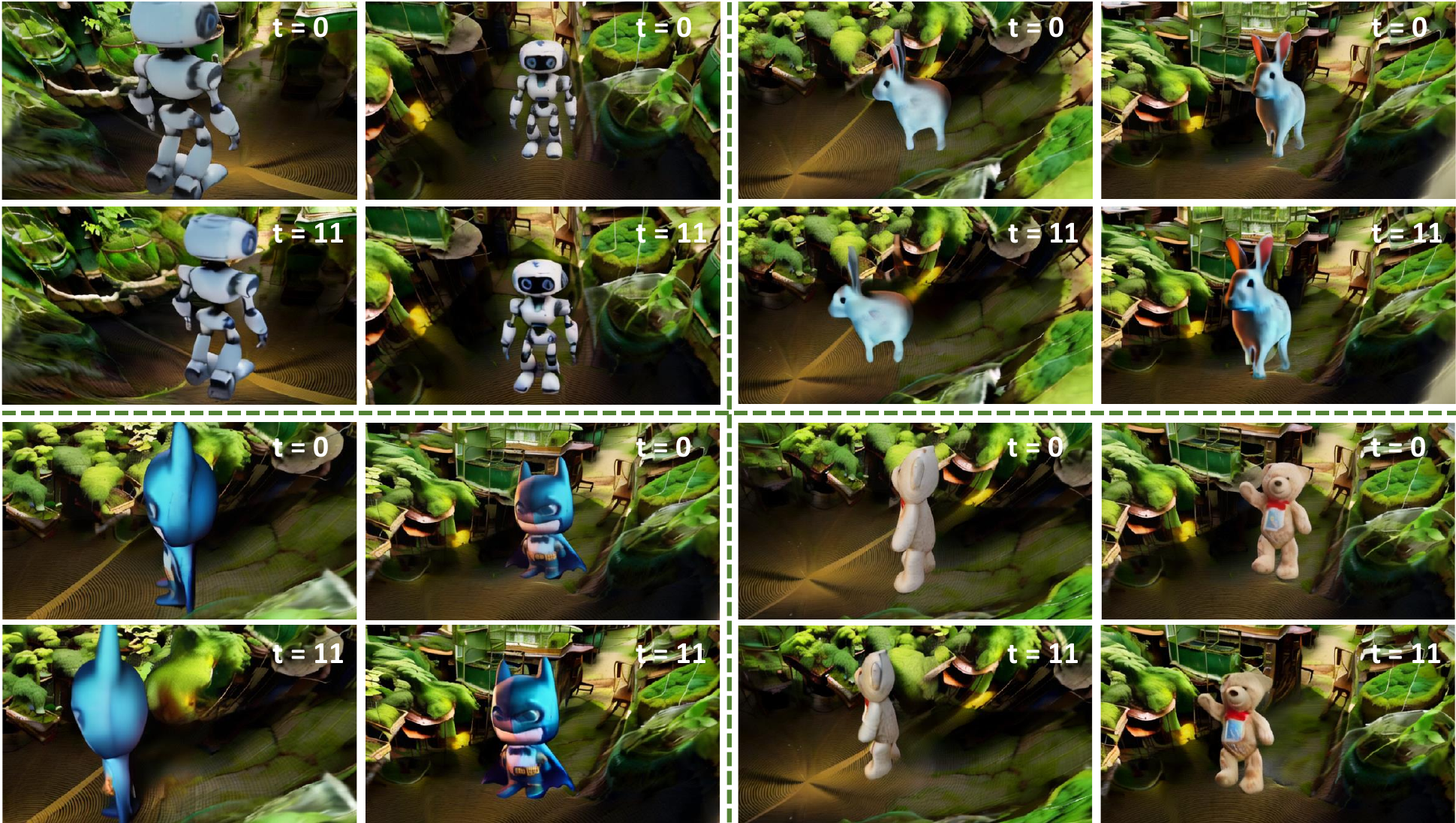}
\caption{A multi-view generated sequence of four different objects moving in the 'greenhouse' scene.}
\label{fig:supple_qual3}
\end{figure*}

\begin{figure*}
\centering
\includegraphics[width=1.0\linewidth]{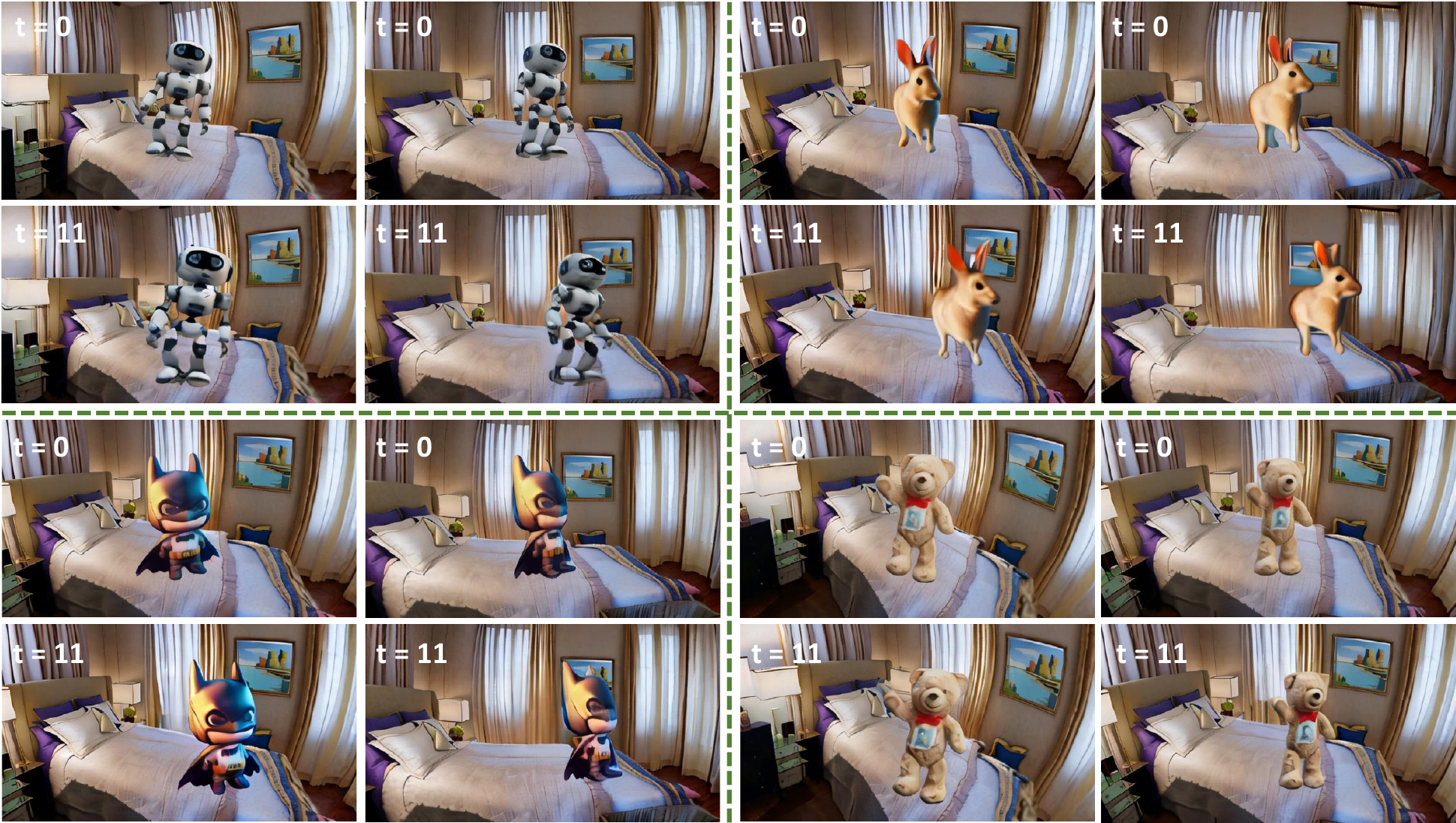}
\caption{A multi-view generated sequence of four different objects jumping in the 'room' scene.}
\label{fig:supple_qual4}
\end{figure*}

\begin{figure*}
\centering
\includegraphics[width=1.0\linewidth]{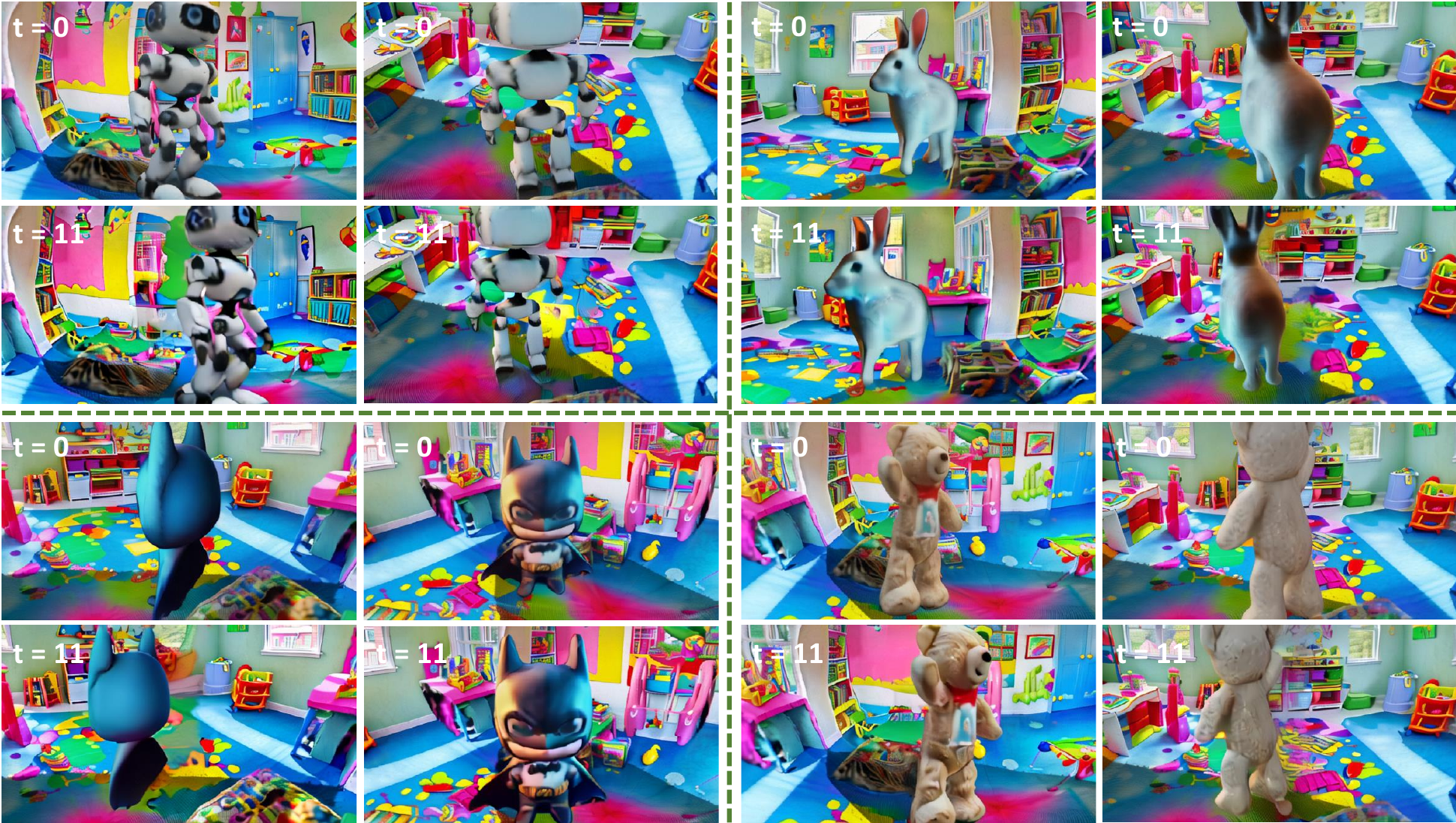}
\caption{A multi-view generated sequence of four different objects moving in the 'playroom' scene.}
\label{fig:supple_qual5}
\end{figure*}

We present additional qualitative comparisons between our final 4D dragged video generated by DragAnything (baseline) and our Local-Global DragAnything (LG-DragAnything) in~\cref{fig:supple_comp1,fig:supple_comp2}. LG-DragAnything successfully models both part-level and global motion, enabling the video diffusion model to move objects more accurately along the input trajectory while minimizing visual artifacts or hallucinations. We also demonstrate that our objects move naturally along the given 3D path within the scene by showing multi-view rendered moving objects in~\cref{fig:supple_qual1,fig:supple_qual2,fig:supple_qual3,fig:supple_qual4,fig:supple_qual5}.

\section{Discussions and Limitations}
\label{sec:supple_discussions}
\begin{figure*}
\centering
\includegraphics[width=1.0\linewidth]{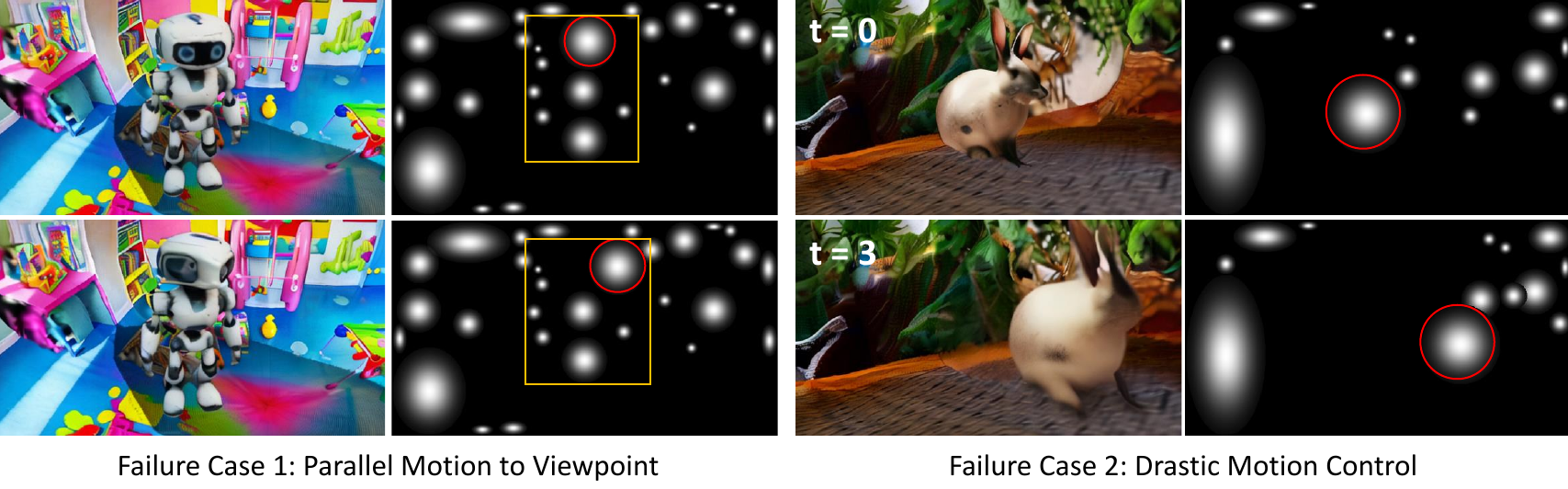}
\caption{\textbf{Failure Cases}: Our method struggles with the motion moving parallel to the viewpoint because of the projected 2D trajectory's ambiguity (Case 1) and drastic motion control (Case 2) which is a common unsolved problem for trajectory-conditioned video generation studies.}
\label{fig:supple_qual6}
\end{figure*}

Our research focuses on 3D object motion control within a scene, therefore, modeling the object's texture under natural scene lighting is beyond the scope of our paper. However, leveraging the provided Drag4D-30 dataset to model realistic object textures within a generated environment under non-Lambertian assumptions presents an interesting direction for future work. In our current work, we identify two interesting failure cases, as illustrated in~\cref{fig:supple_qual6}. First, due to the inherent reliance on 2D trajectories as a condition for our video generation model, motion parallel to the camera view introduces ambiguity (depicted in case 1). For future research, we aim to resolve this issue by introducing a new 3D representation for trajectory conditions. Second, our model faces challenges in handling fast, drastic movements (depicted in case 2). This limitation, commonly observed in recent trajectory-conditioned video generation methods, represents a promising future research direction.

\section{Asset License}
\label{sec:asset}
The licenses of the assets used in the experiments are denoted as follows:

\noindent\textbf{Datasets:}
\begin{itemize}\setlength\itemindent{10pt}
    \item \textbf{VIP-Seg} (Miao et al., 2022): \url{https://github.com/VIPSeg-Dataset/VIPSeg-Dataset}
\end{itemize}

\noindent\textbf{Codes:}
\begin{itemize}\setlength\itemindent{10pt}
    \item \textbf{2D Gaussian Splatting} (Huang et al., 2024): \url{https://github.com/hbb1/2d-gaussian-splatting}
    \item \textbf{InstantMesh} (Xu et al., 2024): \url{https://github.com/TencentARC/InstantMesh}
    \item \textbf{Stable Video Diffusion} (Blattmann et al., 2023): \url{https://huggingface.co/stabilityai/stable-video-diffusion-img2vid-xt}
    \item \textbf{ThreeStudio} (Guo et al., 2023): \url{https://github.com/threestudio-project/threestudio}
    \item \textbf{PanoFusion} (Zhang et al., 2024):
    \url{https://github.com/chengzhag/PanFusion}
\end{itemize}

\end{document}